\documentclass[review]{elsarticle}

\usepackage{lineno,hyperref}
\usepackage{subfigure}
\usepackage{amsthm,amsmath}
\usepackage{amssymb}

\usepackage{makecell,multirow}

\journal{Journal of \LaTeX\ Templates}

\begin{document}

\begin{frontmatter}

\title{Benchmarking Subset Selection from Large Candidate Solution Sets in Evolutionary Multi-objective Optimization}

\author{Ke Shang}
\author{Tianye Shu}
\author{Hisao Ishibuchi}
\cortext[mycorrespondingauthor]{Corresponding author: Hisao Ishibuchi (hisao@sustech.edu.cn).}
\author{Yang Nan}
\author{Lie Meng Pang}

\address{Guangdong Provincial Key Laboratory of Brain-inspired Intelligent Computation, \\Department of Computer Science and Engineering, \\Southern University of Science and Technology, Shenzhen 518055, China}




\begin{abstract}
In the evolutionary multi-objective optimization (EMO) field, the standard practice is to present the final population of an EMO algorithm as the output. However, it has been shown that the final population often includes solutions which are dominated by other solutions generated and discarded in previous generations. Recently, a new EMO framework has been proposed to solve this issue by storing all the non-dominated solutions generated during the evolution in an archive and selecting a subset of solutions from the archive as the output. The key component in this framework is the subset selection from the archive which usually stores a large number of candidate solutions. However, most studies on subset selection focus on small candidate solution sets for environmental selection. There is no benchmark test suite for large-scale subset selection. This paper aims to fill this research gap by proposing a benchmark test suite for subset selection from large candidate solution sets, and comparing some representative methods using the proposed test suite. The proposed test suite together with the benchmarking studies provides a baseline for researchers to understand, use, compare, and develop subset selection methods in the EMO field. 
\end{abstract}


\end{frontmatter}


\section{Introduction}
Subset selection is a popular research topic in the machine learning field. In general, the subset selection problem is to select a subset from a whole set in order to fulfill a specific objective \cite{qian2019distributed}. For example, the influence maximization problem \cite{kempe2003maximizing} involves selecting a subset of users from a social network to maximize the spread of influence. In sparse regression \cite{miller2002subset}, a subset of observation variables is selected in order to achieve the best linear regression model. In feature selection \cite{farahat2011efficient}, a subset of features is selected in order to minimize the reconstruction error of the data matrix. There are also many other subset selection problems such as sensor placement \cite{krause2008near}, maximum coverage \cite{feige1998threshold}, and exemplar-based clustering \cite{dueck2007non}.

The subset selection problem is also involved in the evolutionary multi-objective optimization (EMO) field. For example, in  environmental selection of an EMO algorithm, the next population is selected from the merged population of the current and offspring populations. Subset selection is performed in each generation of an EMO algorithm. Subset selection can be also used as a post-processing procedure after the termination of the execution of an EMO algorithm. In this case, an external archive is maintained to store non-dominated solutions among the examined solutions. Subset selection is performed to select a pre-specified number of solutions from the external archive as the final output. 

Subset selection has been discussed in many studies in the EMO field. The most popular topic is the so-called hypervolume subset selection \cite{bringmann2014two,kuhn2016hypervolume,guerreiro2016greedy}. In this problem, a subset is selected from a candidate solution set in order to maximize the hypervolume \cite{zitzler2003performance} of the selected subset. Another popular topic is the distance-based subset selection \cite{singh2018distance}, which selects a subset in order to maximize the uniformity level of the selected subset \cite{shang2021distance}. There are also some other studies on subset selection based on clustering \cite{chen2021cluster}, reference vectors \cite{deb2013evolutionary}, and other performance indicators such as IGD \cite{coello2004study}, $\varepsilon$+ \cite{zitzler2003performance} and IGD+ \cite{ishibuchi2015modified}. 

\begin{figure}[!htb]
\centering
\subfigure[Standard EMO framework]{                    
\includegraphics[scale=0.3]{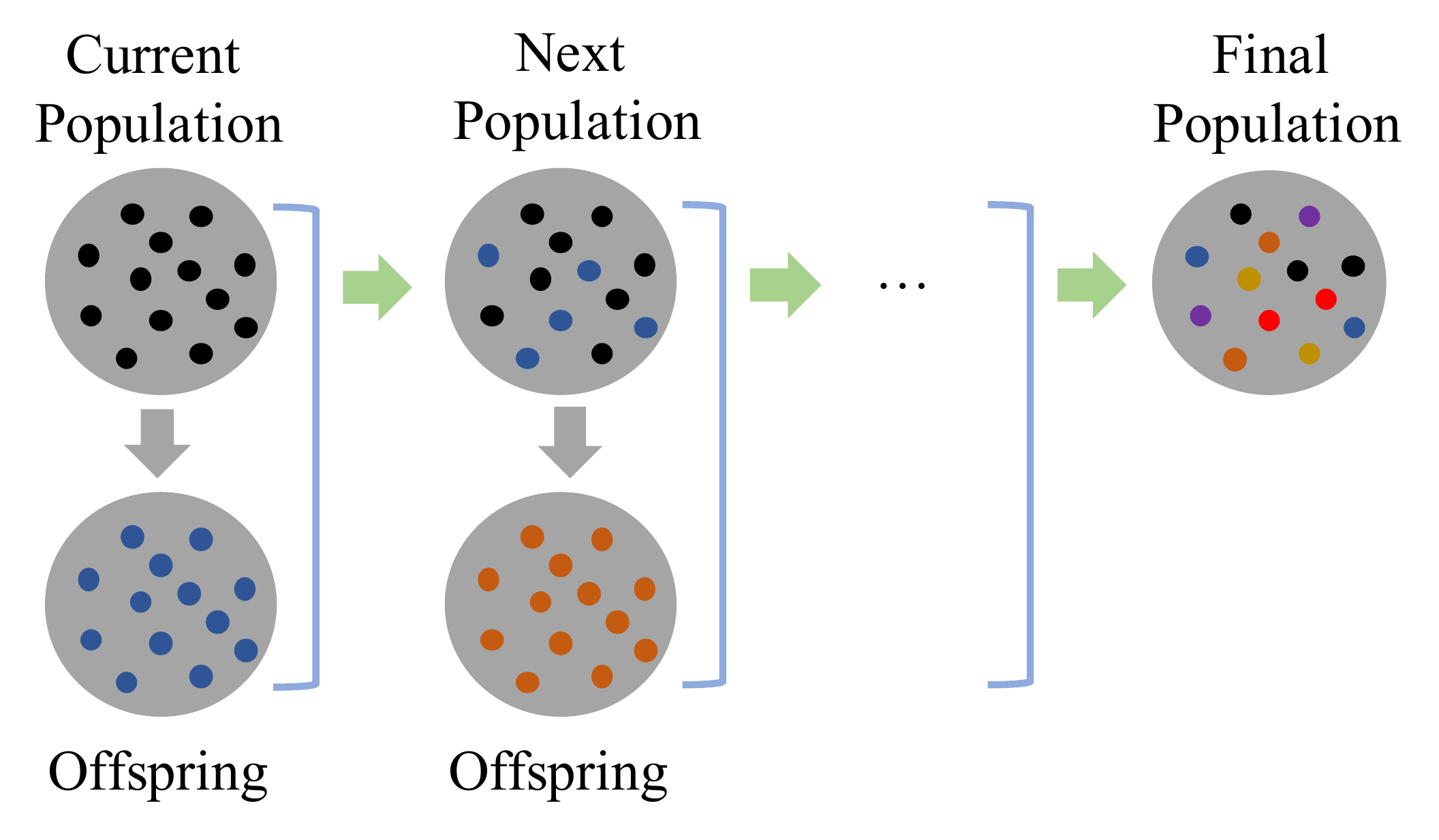}               
}
\subfigure[New EMO framework proposed in \cite{ishibuchi2020new}]{                    
\includegraphics[scale=0.3]{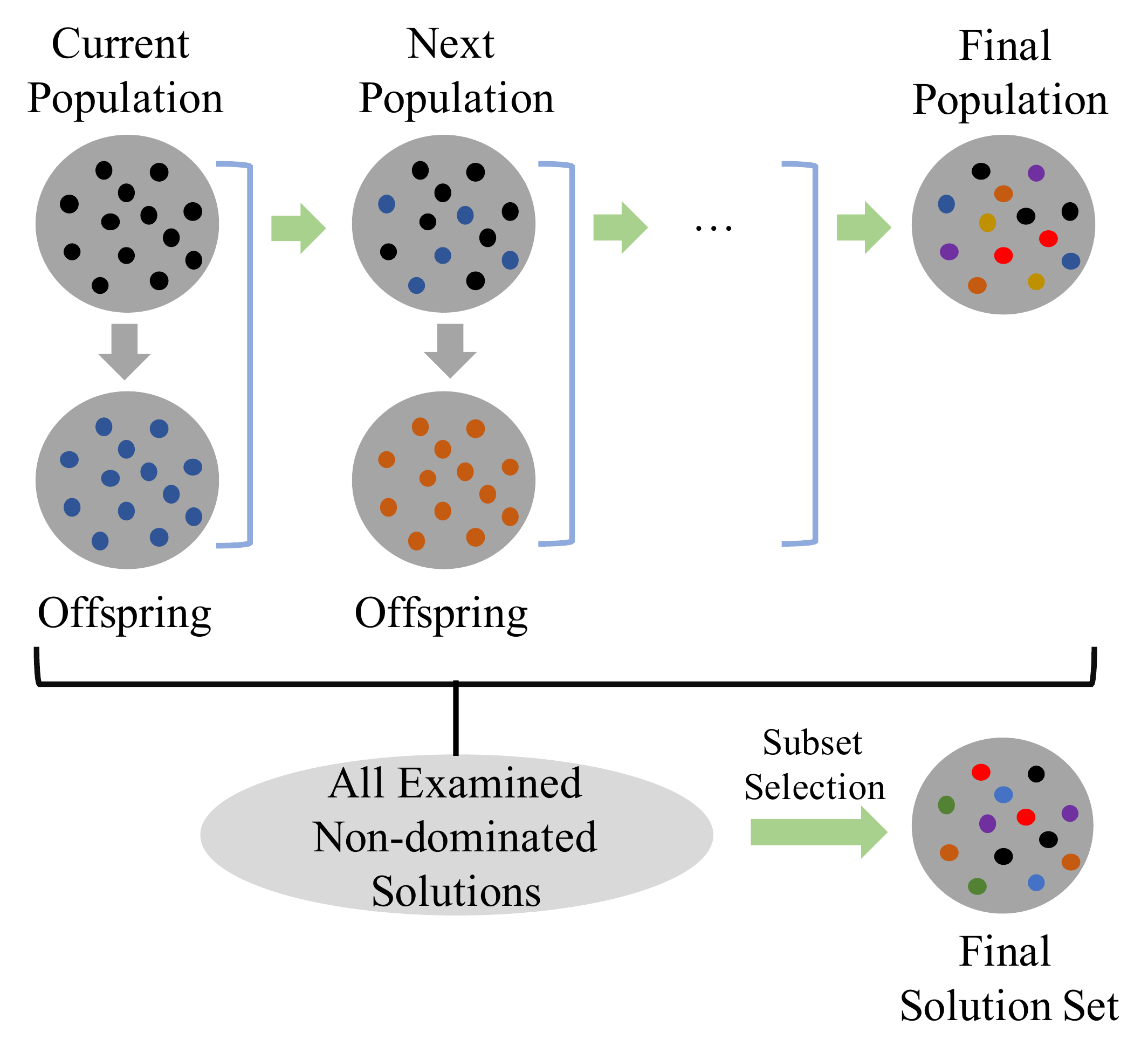}                
}
\caption{An illustration of two EMO frameworks. In (a), the final population is the output of an EMO algorithm. In (b), the selected subset from non-dominated solutions among all the examined solutions is the output of an EMO algorithm.} 
\label{framework}                                                        
\end{figure}

\subsection{Motivation of This Work}
This paper is a benchmarking study on subset selection as a post-processing procedure to select a pre-specified number of solutions from a large number of non-dominated solutions among the examined solutions by an EMO algorithm. The motivation of this paper is elaborated as follows. 
\subsubsection{The drawback of the standard EMO framework}
In most EMO algorithms, the output is the final population as shown in Fig. \ref{framework} (a). However, as shown in \cite{li2019empirical}, the final population is not always a good solution set. In fact, it is often the case for many EMO algorithms that some solutions in the final population are dominated by other solutions generated and discarded during the evolution (i.e., by other solutions which are examined in previous generations but not included in the final population). That is, it is not always a good strategy to present the final population as the output of an EMO algorithm.

In computational experiments in \cite{li2019empirical}, each solution in the final population was checked whether the solution is dominated by any other examined solutions in the execution of an EMO algorithm. Then the percentage of non-dominated solutions in the final population was calculated over 30 runs of each of eight EMO algorithms on each of 37 test problems. From the reported results in \cite{li2019empirical}, we create Table \ref{emo19} where the highest, lowest and average percentages of non-dominated solutions in the final population over the 37 test problems are shown for each of the eight EMO algorithms. We can see from Table \ref{emo19} that in average, the final population of an EMO algorithm contains solutions dominated by some examined solutions. For more details about the experimental results, please refer to \cite{li2019empirical}.


\begin{table}[!tb]
\centering
\fontsize{8.0pt}{0.5\baselineskip}\selectfont
\caption{The percentage of the non-dominated solutions in the final population compared with all the non-dominated solutions generated during the evolution. The higher the better. This table is created from the reported experimental results in Table 1 of \cite{li2019empirical}.}
\begin{tabular*}{\linewidth}{p{3cm}p{2.7cm}p{2.7cm}p{2.7cm}p{1cm}}
\hline
Algorithm & Highest & Lowest & Average  \\\hline
NSGA-II \cite{deb2002fast} &100.0\% &33.6\% &73.54\% \\\hline
NSGA-II+$\epsilon$ \cite{koppen2007substitute} &99.8\% &36.5\% &82.05\% \\\hline
SPEA2 \cite{zitzler2001spea2} &100.0\% & 15.9\%&76.15\% \\\hline
SPEA2+SDE \cite{li2013shift} &100.0\% &60.8\% &91.54\% \\\hline
IBEA \cite{zitzler2004indicator} &99.9\% & 15.9\%&81.62\% \\\hline
SMS-EMOA \cite{beume2007sms} &100.0\% & 99.3\%&99.93\% \\\hline
MOEA/D \cite{zhang2007moea} & 100.0\%& 36.8\%&72.31\% \\\hline
NSGA-III \cite{deb2013evolutionary} &100.0\% & 52.0\%&87.65\% \\\hline
\end{tabular*}
\label{emo19}
\end{table}

\subsubsection{A new EMO framework based on subset selection}
In order to solve the above issue in the standard EMO framework, a new EMO framework is proposed in \cite{ishibuchi2020new}, which is illustrated in Fig. \ref{framework} (b). In the new EMO framework, an unbounded external archive is maintained to store non-dominated solutions among all the examined solutions during the execution of an EMO algorithm. Then, after the termination of the algorithm, a pre-specified number of solutions are selected from the archive and presented as the final output. In this manner, all solutions in the final output are non-dominated solutions among all the examined solutions. Thus, the drawback of the standard EMO framework  can be avoided (when the number of non-dominated solutions is larger than the number of solutions to be presented).

\begin{figure}[!htb]
\centering
\subfigure[Final population (NSGA-II)]{                    
\includegraphics[scale=0.9]{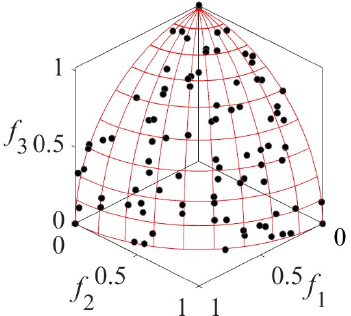}               
\includegraphics[scale=0.9]{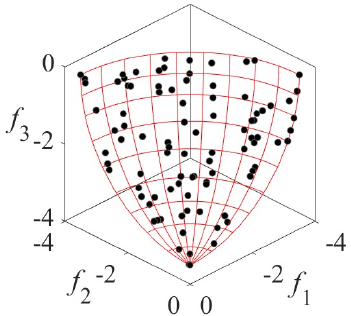}               
}
\subfigure[Selected subset (NSGA-II)]{                    
\includegraphics[scale=0.9]{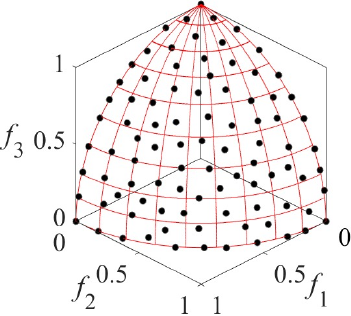}               
\includegraphics[scale=0.9]{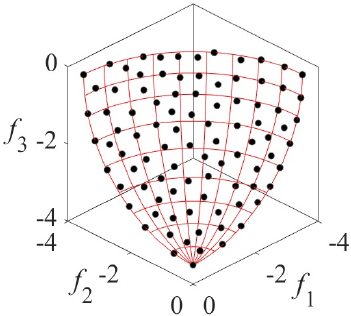}               
}
\subfigure[Final population (MOEA/D-PBI)]{                    
\includegraphics[scale=0.9]{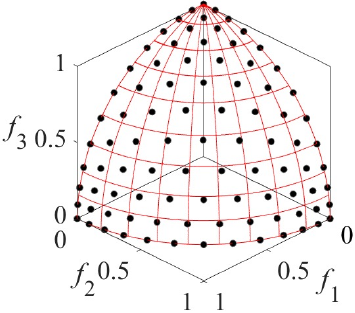}                
\includegraphics[scale=0.9]{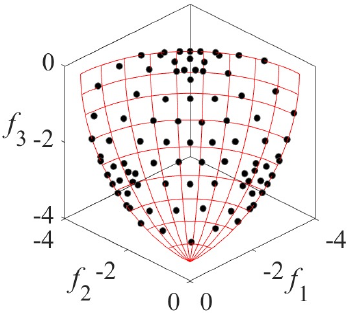}                
}\\
\subfigure[Selected subset (MOEA/D-PBI)]{                    
\includegraphics[scale=0.9]{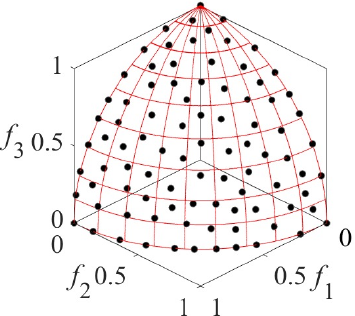}               
\includegraphics[scale=0.9 ]{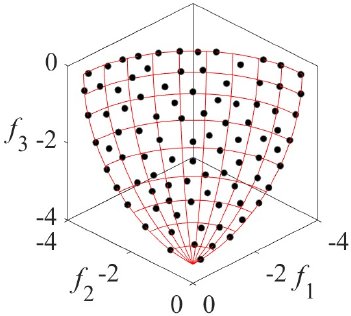}               
}
\caption{Experimental results by NSGA-II and MOEA/D with the standard and the new EMO frameworks on DTLZ2 and MinusDTLZ2 test problems. In the new EMO framework, the distance-based subset selection (DSS) method is used. Figures are from \cite{ishibuchi2020new}.} 
\label{newframework}                                                        
\end{figure}

Moreover, a better distribution of solutions (e.g., more uniformly distributed solutions) can be obtained as the final output of the new EMO framework than the final population of the standard EMO framework. Fig. \ref{newframework} shows the results of NSGA-II and MOEA/D-PBI (i.e., MOEA/D using the PBI function) under the two EMO frameworks on DTLZ2 and MinusDTLZ2 test problems. The distance-based subset selection method \cite{singh2018distance} is used in the new EMO framework. Each algorithm is terminated after the 250th generation with the population size 91 (i.e., after 22,750 solutions are examined).  As shown in Fig. \ref{newframework} (a), the final population of NSGA-II is usually with bad solution distribution. However, by using the new EMO framework as shown in Fig. \ref{newframework} (b), a more uniform solution set can be obtained as the output. For MOEA/D-PBI  in Fig. \ref{newframework} (c), we can see that a uniform solution set can be obtained for the triangular Pareto front whereas a non-uniform solution set is obtained for the inverted triangular Pareto front. By using the new EMO framework in Fig. \ref{newframework} (d), uniform solution sets can be obtained for both types of Pareto fronts. 

\begin{figure}[]
\centering
\subfigure[Selected subset (NSGA-II)]{                    
\includegraphics[scale=0.21]{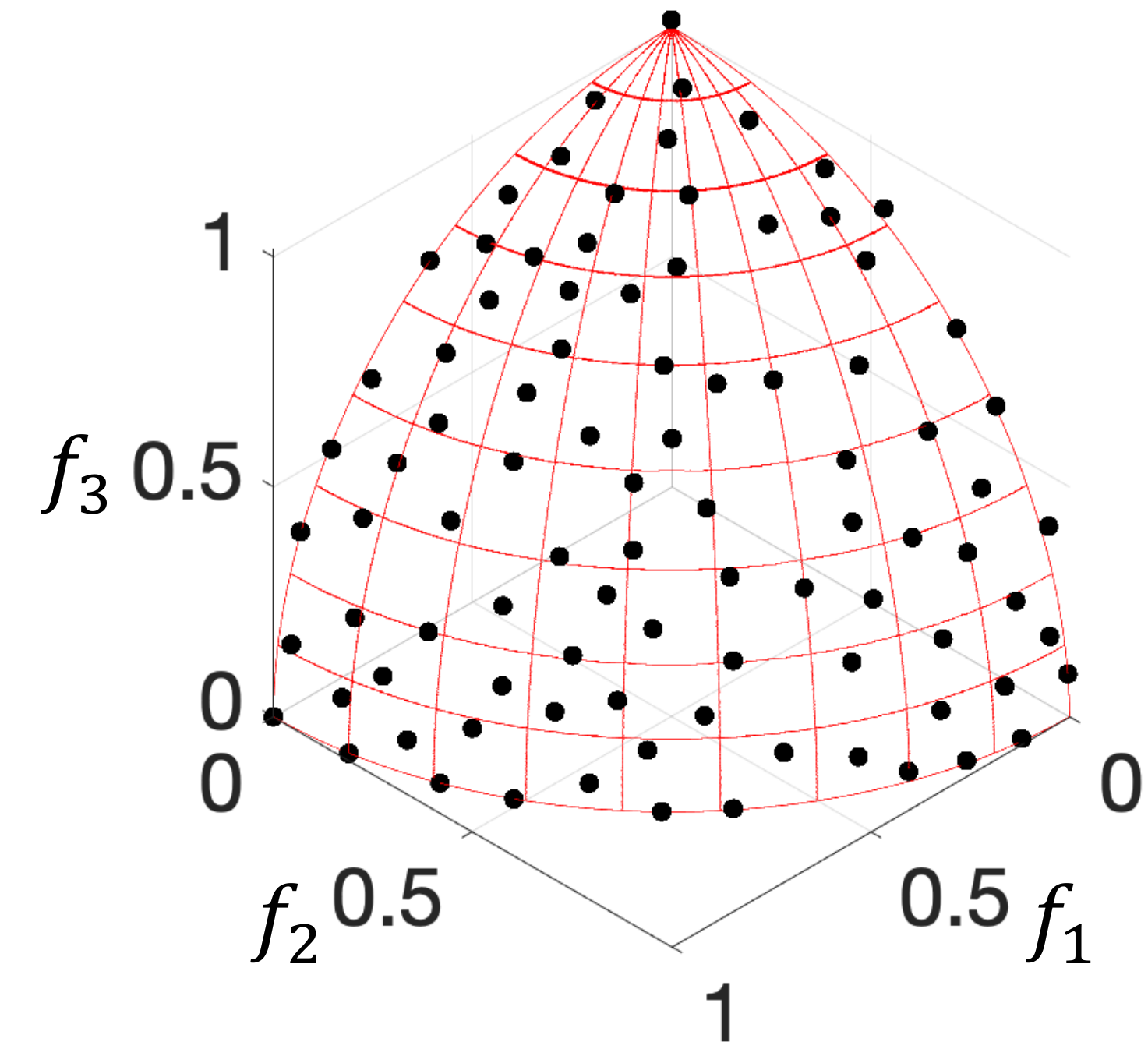}               
\includegraphics[scale=0.21]{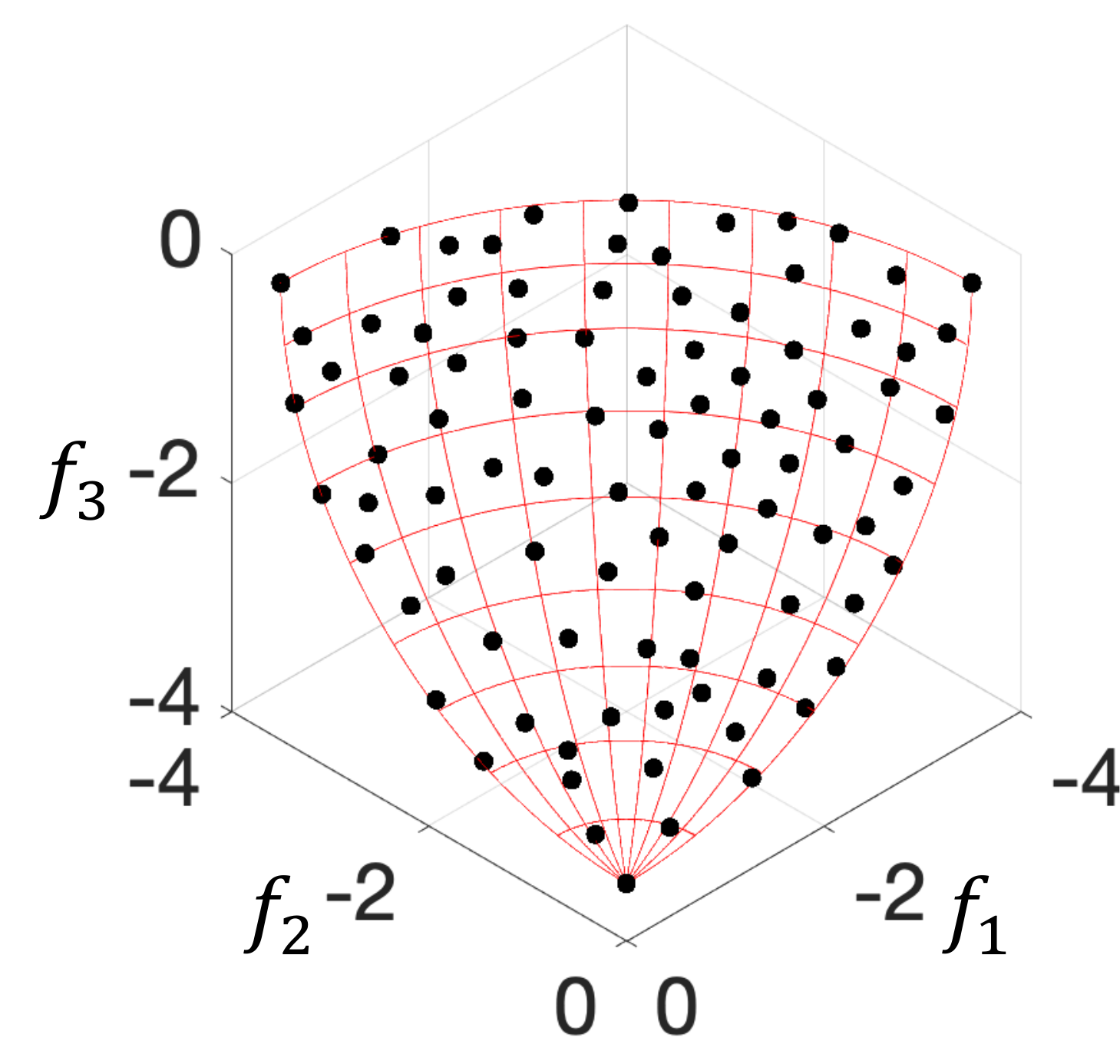}               
}
\caption{Experimental results by NSGA-II with the new EMO framework on DTLZ2 and MinusDTLZ2 test problems. In the new EMO framework,  the unbounded external archive is used only in the last 10 generations, and the distance-based subset selection (DSS) method is used for final solution selection.} 
\label{newnsga2}                                                        
\end{figure}

One may think that it is unrealistic to store all non-dominated solutions in some real-world applications. However, storing solutions is usually much cheaper than examining solutions. If the use of an unbounded external archive throughout the execution of an EMO algorithm is unrealistic for some reason in a specific application problem, we can use it over a realistic number of generations in the final stage of evolution. Fig. \ref{newnsga2} shows the selected solutions after the execution of NSGA-II where an unbounded external archive is used only in the last 10 generations (i.e., for the current population at the 241th generation and the offspring in the (241-250)th generations). Better solution sets are selected in Fig. \ref{newnsga2} than the final population in Fig. 2 (a) whereas Fig. 2 (b) is slightly better than Fig. \ref{newnsga2}. This observation suggests that the subset selection from the archive is useful even when the archive is used only in the final stage of evolution. 

\begin{table}[!tbp]
\centering
\fontsize{8.0pt}{0.5\baselineskip}\selectfont
\caption{The total number of examined solutions in some EMO algorithms.}
\begin{tabular*}{\linewidth}{p{2cm}p{2.5cm}p{2.5cm}p{3cm}p{5cm}}
\hline
Algorithm & Population Size & Generations & Examined Solutions  \\\hline
NSGA-III \cite{deb2013evolutionary} &92, 212, 156, 276, and 136 for 3, 5, 8, 10, and 15 objectives, respectively & 600, 1000, 1250, 2000, and 3000 for 3, 5, 8, 10, and 15 objectives, respectively&55,200, 212,000, 195,000, 552,000, and 408,000 for  3, 5, 8, 10, and 15 objectives, respectively        \\\hline
FDEA-I, FDEA-II \cite{qiu2021evolutionary}&212, 156, 275, and 136 for 5, 8, 10, and 15 objectives, respectively&-&150,000 for WFG3 and MaF3, and 100,000 for the other problems \\\hline
SRV \cite{liu2020self} &153, 165, 210, 238, and 275 for 3, 4, 5, 7, and 10 objectives, respectively & 600, 700, 800, 900, and 1000 for 3, 4, 5, 7, and 10 objectives, respectively& 91,800, 115,500, 168,000, 214,200, and 275,000 for 3, 4, 5, 7, and 10 objectives, respectively \\\hline
MultiGPO \cite{zhu2021new}&210, 240, 275, 240, and 210 for 5, 8, 10, 15, and 20 objectives, respectively&500 (MaF problems)&105,000, 120,000, 137,500, 120,000, and 105,000 for 5, 8, 10, 15, and 20 objectives, respectively\\\hline
C-TAEA \cite{li2018two}&100 (Water Distribution Network Optimization)&10,000&1,000,000\\\hline
MOEA/D-IFM \cite{zhang2020enhancing}&100&10,000&1,000,000\\\hline
LMEA \cite{zhang2016decision}&105, 126,  and 275 for 3, 5, and 10 objectives, respectively& - &1,000,000,   6,800,000,   17,000,000,  50,000,000, and 230,000,000 for each test instance with 100, 500, 1000, 2000, and 5000 decision variables, respectively\\\hline
\end{tabular*}
\label{examinedsol}
\end{table}

\subsubsection{Subset selection from large candidate solution sets}
In the new EMO framework, the key component is the subset selection from the archive. Since the archive stores non-dominated solutions among the examined solutions during the execution of an EMO algorithm, usually it contains a huge number of non-dominated solutions. For example, in many-objective optimization, the total number of examined solutions in a single run of an EMO algorithm is often hundreds of thousands (100K - 600K). In large-scale many-objective optimization and constrained multi-objective optimization, the total number of examined solutions in a single run of an EMO algorithm is often one million and more. Table \ref{examinedsol} shows different specifications about the number of examined solutions in a single run of an EMO algorithm (i.e., termination condition) in some recent studies. We can see that more than 100K solutions are examined in almost all cases, and one million or more solutions are examined in some studies.

From the above analysis, we can see that in order to use the new EMO framework, an efficient subset selection method is needed in order to select the final solution set from a large candidate solution set.  However, most studies on subset selection are for environmental selection where the candidate solution set is small \cite{beume2007sms,jiang2015simple,deb2013evolutionary}. When subset selection is used as a post-processing procedure to choose a final solution set, most studies are for two- and three-objective problems (i.e., subset selection is in low-dimensional objective spaces \cite{bringmann2014two,kuhn2016hypervolume,bringmann2017maximum,guerreiro2016greedy}). Thus, there is a research gap on subset selection from large candidate solution sets with many objectives. This motivates us to set up a baseline for this research topic. Therefore, in this paper, we first propose a benchmark test suite for subset selection from large candidate solution sets. After that, we conduct benchmarking studies by comparing several existing subset selection methods on the proposed test suite.

The main aim of this paper is to provide EMO researchers with a number of ready-to-use candidate solution sets for subset selection and experimental results by representative subset selection methods. This is to encourage subset selection research in the EMO field, which will lead to the improvement of the final outputs from existing EMO algorithms and the development of new EMO algorithms.  





\subsection{Organization of This Paper}
The rest of this paper is organized as follows. Section II explains subset selection in evolutionary multi-objective optimization. Section III proposes a benchmark test suite for subset selection. Section IV conducts benchmarking studies on some representative subset selection methods using the proposed test suite. Finally, Section V concludes the paper.

\section{Subset Selection in Evolutionary Multi-objective Optimization}
In this section, we explain subset selection in evolutionary multi-objective optimization (EMO). First, we give a formal mathematical definition for subset selection in EMO. Then, we briefly review existing subset selection methods in the EMO field by categorizing them into different types.

\subsection{Mathematical Definition}
In the EMO field, the subset selection task is to select a pre-specified number of solutions from a non-dominated candidate solution set in order to fulfill some specific objective. Formally, given a non-dominated solution set $A$ with $|A|=n$, an objective function $f: 2^A\rightarrow \mathbb{R}$, the subset selection problem aims to find a subset $S\subset A$ with $|S|=k$ so that
\begin{equation}
\label{sseq}
S = {\arg\max}_{S'\subset A,|S'|=k} f(S').
\end{equation}

Without loss of generality, we assume $f$ in \eqref{sseq} is to be maximized. For example, $f$ can be the hypervolume indicator. If we want to use the IGD indicator which is usually to be minimized, we can simply add a minus sign to the IGD indicator to get a maximization problem as in \eqref{sseq}. 

In this paper, we consider subset selection in the objective space. Thus, for an $m$-objective problem, a solution means an $m$-dimensional vector (i.e., a point in the objective space).

\subsection{Subset Selection Methods}
\label{ss-methods}
Subset selection methods can be classified into the following five categories:
\subsubsection{Indicator-based subset selection}
This type is the most widely investigated one. The objective function $f$ of the subset selection problem in (1) is based on a performance indicator such as hypervolume \cite{zitzler2003performance}, IGD \cite{coello2004study}, IGD+ \cite{ishibuchi2015modified} and $\varepsilon$+ \cite{zitzler2003performance}. In Table \ref{ssmethod}, indicator-based subset selection methods are divided into four classes (i.e., HSS, IGDSS, IGD+SS and e+SS) depending on the performance indicator used for subset selection. The objective function corresponding to each indicator is explained in Table \ref{ssmethod}.

Different types of optimization algorithms have been used for subset selection based on each indicator. For example, HSS has been addressed using three types of optimization algorithms: exact algorithms, greedy algorithms, and evolutionary algorithms \cite{shang2020survey}.  Exact algorithms \cite{bringmann2014two,kuhn2016hypervolume} aim to find the optimal subset. However, this type of algorithms are time-consuming due to the NP-hardness of the HSS problem \cite{bringmann2017maximum}.  Greedy algorithms \cite{bradstreet2007incrementally,guerreiro2016greedy} and evolutionary algorithms \cite{ishibuchi2009selecting,friedrich2014maximizing} are more efficient and practical for subset selection from large candidate solution sets. In general, greedy algorithms are more widely investigated and applied in the EMO field.

\begin{table}[!tb]
\centering
\fontsize{8.0pt}{0.5\baselineskip}\selectfont
\caption{The subset selection methods in the EMO field, and their corresponding objective functions in \eqref{sseq}.}
\begin{tabular*}{\linewidth}{p{2cm}p{9cm}p{2.5cm}p{3cm}p{5cm}}
\hline
Method & Objective function $f$ in \eqref{sseq}   \\\hline
HSS & $f(S) = \mathcal{L}\left(\bigcup_{\mathbf{s}\in S}\left\{\mathbf{s}'|\mathbf{s}\prec \mathbf{s}'\prec \mathbf{r}\right\}\right)$,
where $\mathcal{L}(.)$ is the Lebesgue measure of a set, $\mathbf{r}$ is a reference point, and $\mathbf{s}\prec \mathbf{s}'$ denotes that $\mathbf{s}$ Pareto dominates $\mathbf{s}'$. \\\hline
IGDSS &  $f(S) = -\frac{1}{|R|}\sum_{\mathbf{r}\in R}\min_{\mathbf{s}\in S}d(\mathbf{s},\mathbf{r})$, where $R$ is a reference point set, $d$ is the Euclidean distance. \\\hline
IGD+SS &$f(S) = -\frac{1}{|R|}\sum_{\mathbf{r}\in R}\min_{\mathbf{s}\in S}d^+(\mathbf{s},\mathbf{r})$, where $R$ is a reference point set, $d^+$ is the IGD+ distance which is defined as $d^+(\mathbf{s},\mathbf{r}) = \sqrt{\sum_{i=1}^m(\max\{0,s_i-r_i\})^2}$ for minimization problems. \\\hline
$\varepsilon$+SS & $f(S) = -\max_{\mathbf{r}\in R}\min_{\mathbf{s}\in S}\max_{i=1}^m\{s_i-r_i\}$, where $R$ is a reference point set.  \\\hline
DSS  & $f(S) =  \min_{\mathbf{x},\mathbf{y}\in S, \mathbf{x}\neq \mathbf{y}}d(\mathbf{x},\mathbf{y})$, where $d$ is the Euclidean distance.
 \\\hline
CSS & Based on k-means: $f(S) = -\sum_{\mathbf{r}\in R}\min_{\mathbf{s}\in S}d(\mathbf{s},\mathbf{r})^2$. Based on k-medoids: $f(S) = -\sum_{\mathbf{r}\in R}\min_{\mathbf{s}\in S}d(\mathbf{s},\mathbf{r})$, where $R$ is a reference point set, and $d$ is the Euclidean distance.\\\hline
RVSS & $f(S) = -\sum_{\mathbf{v}\in V}\min_{\mathbf{s}\in S}d(\mathbf{s},\mathbf{v})$, where $V$ is a reference vector set with $|V| = k$, $d$ is a distance metric which can be the perpendicular distance, i.e., $d(\mathbf{s},\mathbf{v})=\left\|(\mathbf{s}-\mathbf{v}^\mathrm{T}\mathbf{s}\mathbf{v}/\left\|\mathbf{v}\right\|^2)\right\|$, or the angle distance, i.e., $d(\mathbf{s},\mathbf{v})=1-\frac{\mathbf{s}\mathbf{v}^\mathrm{T}}{\sqrt{(\mathbf{s}\mathbf{s}^\mathrm{T})(\mathbf{v}\mathbf{v}^\mathrm{T})}}$.\\\hline
\end{tabular*}
\label{ssmethod}
\end{table}

\subsubsection{Distance-based subset selection}
This method is proposed in \cite{singh2018distance}. Solutions are selected one by one from the candidate solution set to find a subset of a pre-specified size (i.e., with $k$ solutions). Initially the subset is empty. In each step, the solution which is farthest to the current subset is selected from the candidate solution set and added into the subset. This procedure repeats until $k$ solutions are selected.

As proved in \cite{shang2021distance}, the distance-based subset selection is a greedy algorithm in nature with respect to maximizing the uniformity level \cite{sayin2000measuring} of the subset (which is defined as the minimum distance between two solutions in the subset). Based on this conclusion, we can see that the distance-based subset selection can be viewed as an indicator-based method where the performance indicator is the uniformity level. The sixth row of Table \ref{ssmethod} shows the objective function for the distance-based subset selection (denoted as DSS).

\subsubsection{Clustering-based subset selection}
Clustering is a popular research topic for data analysis in the machine learning field where it is also used for subset selection \cite{daszykowski2002representative}. In \cite{chen2021cluster}, clustering methods are applied to subset selection in the EMO field. The general idea of clustering-based subset selection is to use a clustering method (e.g., k-means and k-medoids) to classify the non-dominated solutions in the candidate solution set into $k$ different clusters. Then a representative solution is selected from each cluster to construct a subset with $k$ solutions. The seventh row in Table \ref{ssmethod} shows the objective functions for the clustering-based subset selection (denoted as CSS) based on k-means and k-medoids.

As shown in \cite{chen2021cluster}, CSS based on k-medoids has the same objective function as IGDSS. We can also see that CSS based on k-means has a similar objective function to IGDSS. Thus, CSS and IGDSS are closely related to each other.

\subsubsection{Reference vector-based subset selection}
This type of subset selection selects solutions using a set of pre-specified reference vectors. The same idea  is used for environmental selection in some well-known EMO algorithms such as MOEA/D \cite{zhang2007moea} and NSGA-III \cite{deb2013evolutionary}. First, a set of reference vectors is generated. The Das and Dennis method \cite{das1998normal} is widely-used for this purpose. Then, the solution with the minimum distance to each reference vector is selected from the candidate solution set. Usually, the perpendicular distance or the angle distance is used as the distance measure between a solution and a reference vector. In reference vector-based subset selection, the objective function of the subset selection problem can be viewed as the sum of the difference between each reference vector and the nearest solution in the selected subset. The last row in Table \ref{ssmethod}  explains the reference vector-based subset selection (denoted as RVSS) in the framework of the subset selection problem with the objective function $f$.

One important issue in this type of subset selection is the specification of reference vectors since the selected subset directly depends on their distribution. In general, the Das and Dennis method generates a reference vector set with a triangular shape. A uniform subset can be selected when the candidate solution set is on a triangular Pareto front. However, it is not easy to select a uniform subset if the candidate solution set is on an irregular Pareto front (e.g., inverted triangular Pareto front \cite{ishibuchi2017performance}). 


\subsubsection{Other subset selection methods}
In principle, any environmental selection mechanism in EMO algorithms can be used for subset selection. In general, an EMO algorithm aims to find a solution set with good convergence and good diversity. Since all candidate solutions are non-dominated in our context, any  diversity maintenance mechanism of an EMO algorithm can be used for subset selection (without worrying about the convergence of the selected subset). For example, it is possible to use the crowding distance mechanism in NSGA-II for subset selection. However, it is often the case that environmental selection mechanisms in EMO algorithms have no clear objective functions when they are viewed as subset selection methods. If the decision maker wants to know the reason why the presented subset is selected, it may be important to use a clear objective function in subset selection.

\subsection{Discussions}
It should be noted that the subset selection problem is often NP-hard. Complete enumeration takes $\binom{n}{k}$ time and has a time complexity of $O(2^{\min\{k, n-k\}})$. Special cases of subset selection offer a much faster solution, especially if $k$ (the size of the subset) is close to nor close to 0. Moreover, many subset selection problems can be approximated by greedy algorithms. For instance, subset selection for submodular set function. It has been proved that the hypervolume, IGD, and IGD+ indicators are submodular set functions \cite{ulrich2012bounding,chen2021fast}. The greedy algorithm can provide a $(1-1/e)$ approximation guarantee \cite{nemhauser1978analysis}.

\section{Proposal of A Benchmark Test Suite}
To the best of our knowledge,  there is no benchmark test suite for the subset selection research in the EMO community. Each study on subset selection uses its own candidate  solution sets. In this section, we create a benchmark test suite for subset selection. 

We create a test suite (i.e., a set of non-dominated candidate solution sets) using two methods. One method is to create candidate solution sets by directly sampling points on the Pareto fronts. The other method is to use  EMO algorithms on some test problems. Candidate  solution sets created by these two methods are explained in the following subsections. 

\subsection{Candidate Solution Sets on Pareto Fronts}
\begin{figure}[!htb]
\centering
\subfigure[]{                    
\includegraphics[scale=0.22]{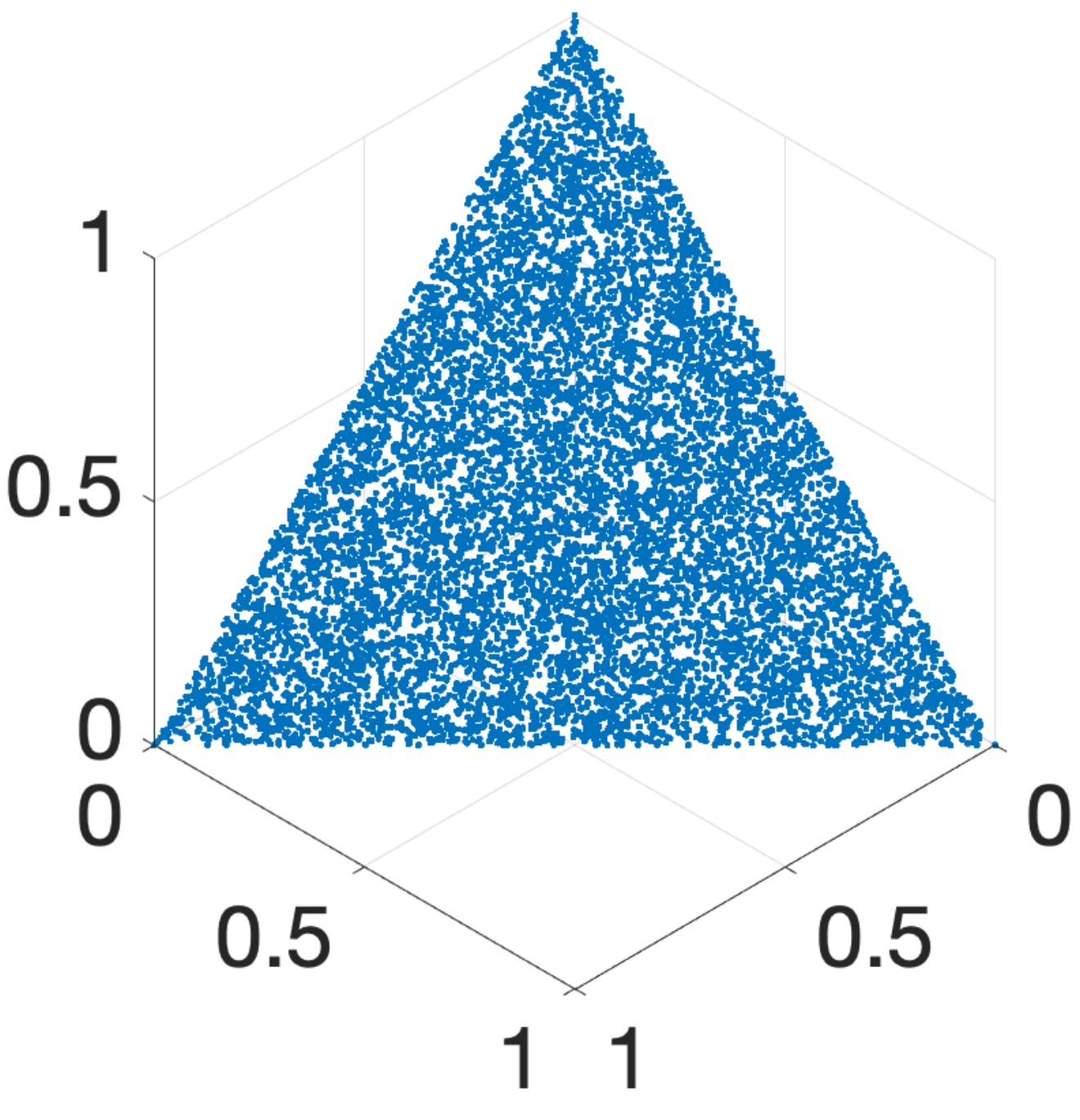}               
}
\subfigure[]{                    
\includegraphics[scale=0.22]{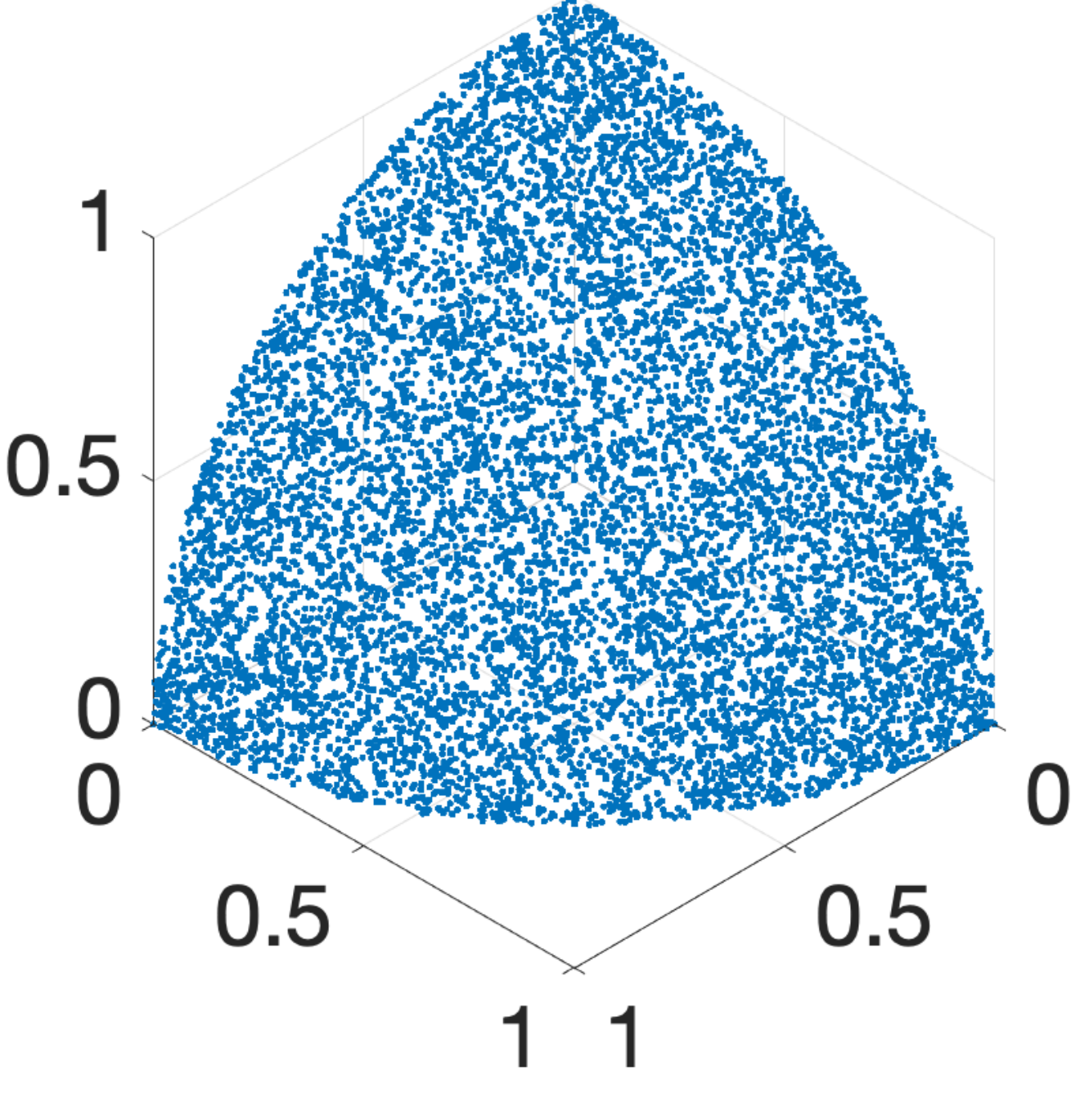}                
}
\subfigure[]{                    
\includegraphics[scale=0.22]{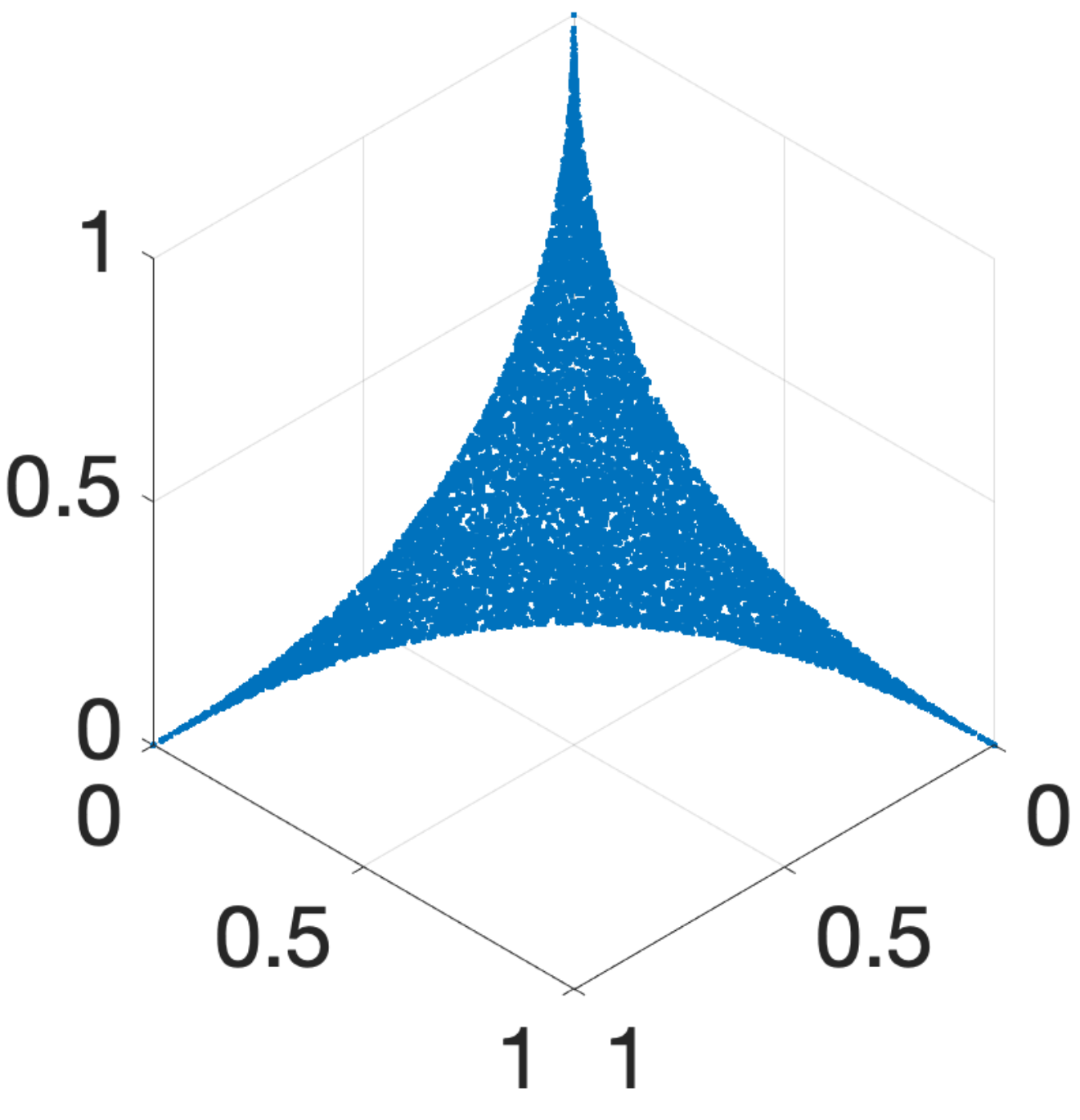}                
}
\subfigure[]{                  
\includegraphics[scale=0.22]{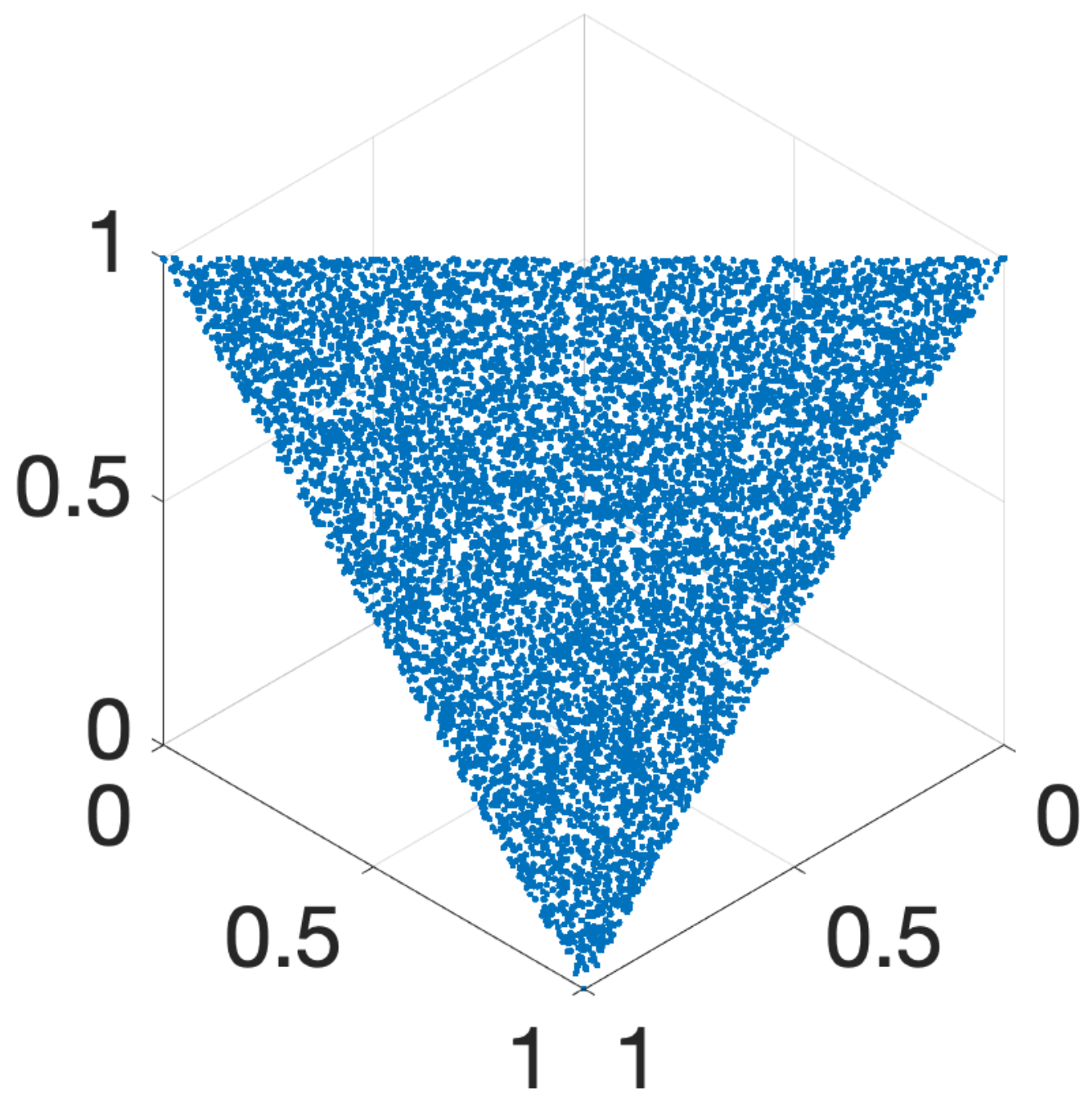}                
}
\subfigure[]{                    
\includegraphics[scale=0.22]{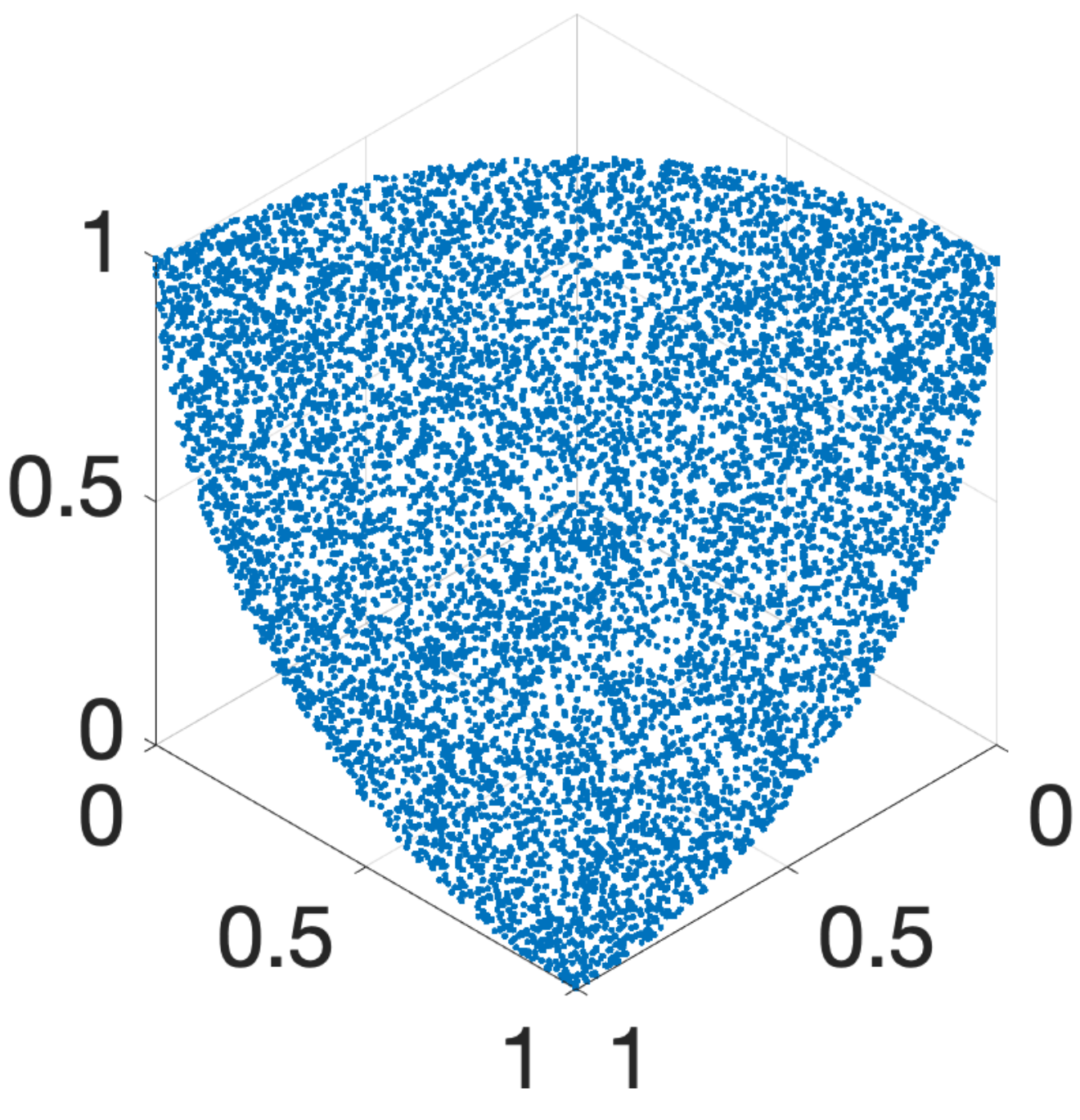}                
}
\subfigure[]{                    
\includegraphics[scale=0.22]{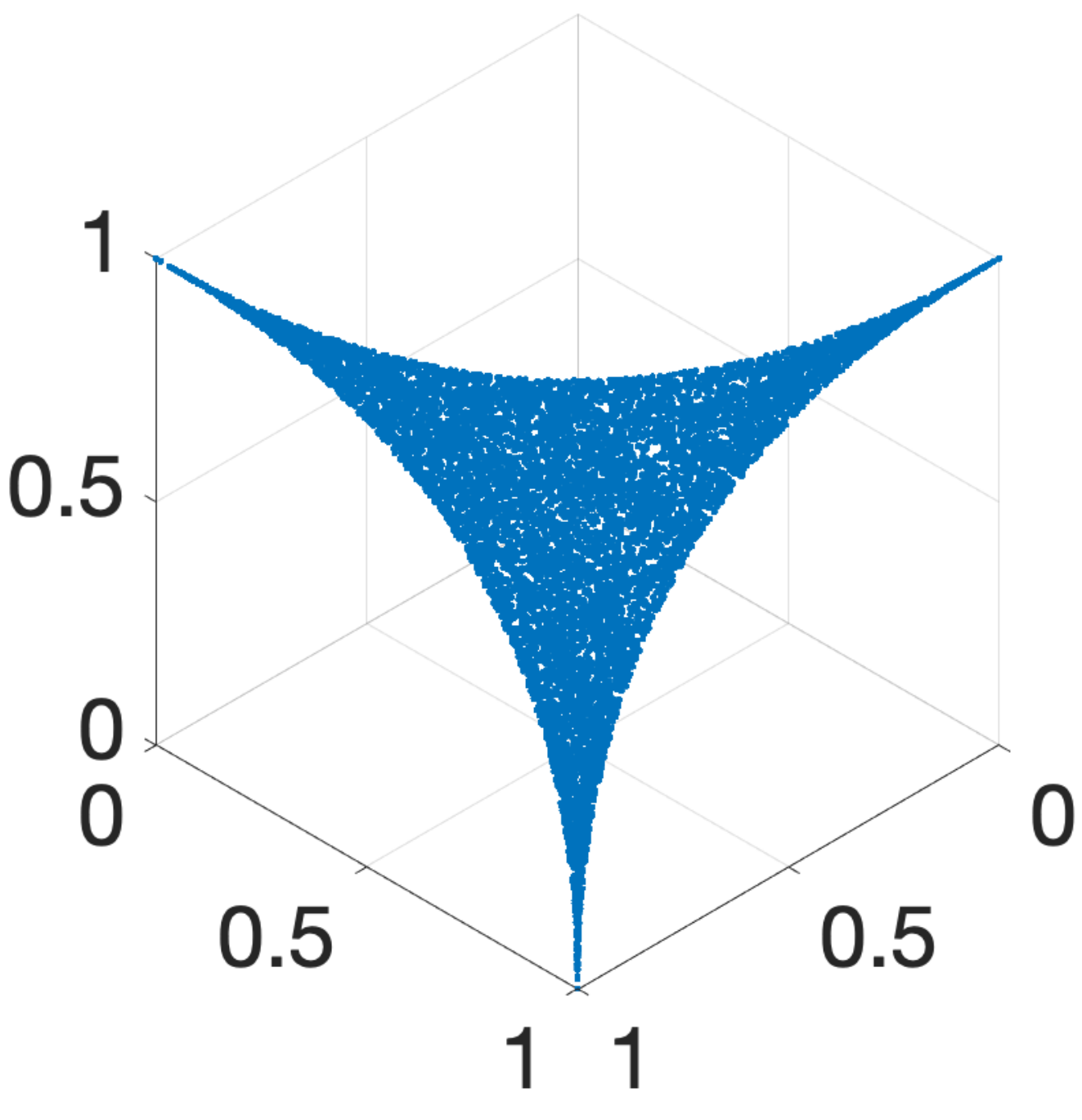}                
}
\caption{An illustration of the six types of Pareto fronts in a three-objective space. (a) Linear Triangular. (b) Concave Triangular. (c) Convex Triangular. (d) Linear Inverted Triangular. (e) Convex Inverted Triangular. (f) Concave Inverted Triangular.} 
\label{sixpf1}                                                        
\end{figure}

The first type of candidate solution sets is created by directly sampling points on different Pareto fronts. We consider six types of Pareto fronts: linear, convex, and concave Pareto fronts with triangular and inverted triangular shapes. The detailed formulations of the six types of Pareto fronts are summarized in Table \ref{sixpf}. Fig. \ref{sixpf1} shows the six Pareto fronts in a three-objective space (i.e., when $m=3$).

\begin{table}[!htb]
\centering
\fontsize{8.0pt}{0.5\baselineskip}\selectfont
\caption{Six types of Pareto fronts considered for generating candidate solution sets.}
\begin{tabular}{l|c}
\hline
Pareto front & Formulation  \\\hline
Linear Triangular&$\sum_{i=1}^m f_i = 1$ and $f_i\geq 0$\\\hline
Convex Triangular&$\sum_{i=1}^m \sqrt{f_i} = 1$ and $f_i\geq 0$ \\\hline
Concave Triangular&$\sum_{i=1}^m f_i^2=1$ and $f_i\geq 0$ \\\hline
Linear Inverted Triangular&$\sum_{i=1}^m (1-f_i) = 1$ and $0\leq f_i\leq 1$ \\\hline
Convex Inverted Triangular&$\sum_{i=1}^m (1-f_i)^2=1$ and $0\leq f_i\leq 1$\\\hline
Concave Inverted Triangular&$\sum_{i=1}^m \sqrt{1-f_i} = 1$ and $0\leq f_i\leq 1$ \\\hline
\end{tabular}
\label{sixpf}
\end{table}

From each of the six Pareto fronts, we randomly sample 10K, 100K and 1M solutions. The number of objectives (i.e., $m$) is set as 3, 5, 8, and 10. As a result, we have 6 (Pareto fronts) $\times$ 3 (specifications of the number of solutions) $\times$ 4 (specifications of the number of objectives) $=$ 72 candidate solution sets. 

We use the method in \cite{ahmadi2019uniform} to uniformly sample points on each Pareto front. For the three triangular Pareto fronts (which are unit $l_p$ spheres with $p\in\{1,2,0.5\}$), the  sampling method in \cite{ahmadi2019uniform}\footnote{Source code available at \url{http://bit.ly/JSPI-GLp}.} is described as follows. 1) Randomly sample point $\mathbf{x}\in \mathbb{R}^m$ according to the exponential power distribution with density $f(x) = \frac{1}{2\Gamma(1+1/p)}e^{-|x|^p}$ where $\Gamma(t) = \int_{0}^{\infty}x^{t-1}e^{-x}dx$ is the Gamma function. 2) Obtain the sampled point $\mathbf{s} = |\mathbf{x}|/\left \| \mathbf{x} \right \|_p$. The pseudocode of the sampling procedure is provided in Section I of the supplementary material.

For the inverted triangular Pareto fronts, we first sample the candidate solution sets on the corresponding triangular Pareto fronts (i.e., we use the sampled candidate solution sets for the triangular Pareto fronts). Then we apply a rotation transformation to obtain desired candidate solution sets. Fig. \ref{sixpf1} shows the sampled candidate solution sets with 10,000 solutions on the six Pareto fronts for $m=3$.

\subsection{Candidate Solution Sets from EMO Algorithms}
The second type of candidate solution sets is created by running EMO algorithms on different multi-objective test problems. We consider the following six popular test problems: DTLZ1 \cite{deb2005scalable}, DTLZ2 \cite{deb2005scalable}, MinusDTLZ1 \cite{ishibuchi2017performance}, MinusDTLZ2 \cite{ishibuchi2017performance}, DTLZ7 \cite{deb2005scalable}, and WFG3 \cite{huband2006review}. The number of objectives $m$ is set as 3, 5, 8, and 10. We run three representative EMO algorithms (i.e., NSGA-II, NSGA-III, and MOEA/D-PBI, i.e., MOEA/D with the PBI function) with the standard setting of the population size (i.e., 91, 210, 156, and 275 for 3, 5, 8, and 10 objectives, respectively) on the six test problems, and create non-dominated solution sets after 10K, 100K and 1M solution evaluations. Each candidate solution set contains all non-dominated solutions among the examined solutions in each setting. As a result, we have 6 (test problems) $\times$ 4 (specifications of the number of objectives) $\times$ 3 (EMO algorithms) $\times$ 3 (termination conditions, i.e., specifications of the number of examined solutions) $=$ 216 candidate solution sets.

Some statistical analyses on the generated candidate solution sets are provided in Section II of the supplementary material.  Interested readers can refer to the analyses to more deeply understand the candidate solution sets.

\subsection{Suggested Candidate Solution Sets}
Using the above-mentioned methods for candidate solution set generation, we can generate a total number of 288 candidate solution sets (72 solution sets generated on the Pareto fronts and 216 solution sets generated by the EMO algorithms), which is our whole  benchmark test suite. 

Now we suggest a small number of candidate solution sets from these 288 solution sets as a small benchmark test suite for subset selection. For the candidate solution sets on Pareto fronts, we suggest the 5- and 10-objective candidate solution sets with 100K and 1M solutions (since these candidate solution sets are more challenging than those with less solutions and/or less objectives). Therefore, there are 24 candidate solution sets in total. For the candidate solution sets from the EMO algorithms, we suggest the 5- and 10-objective candidate solution sets obtained by NSGA-III with 100K and 1M function evaluations (since NSGA-III has been frequently used for many-objective problems). Therefore, there are 24 candidate solution sets in total. Thus, we have 48 candidate solution sets in the small benchmark test suite. 

Ideally, it is advisable to use the whole benchmark test suite for subset selection research since it includes a wide variety of candidate solution sets. However, the reduced benchmark test suite can be useful when the use of the whole benchmark test suite is not realistic for some reason (e.g., when one wants to evaluate subset selection methods in a limited time using a limited computational resource). 

\section{Benchmarking Studies}
In this section, we conduct benchmarking studies on some representative subset selection methods using the proposed benchmark test suite. First, we describe the experimental settings for the benchmarking studies. Then, we present the results and analysis.


\subsection{Experimental Settings}
\subsubsection{Compared methods}
We consider the following 10 subset selection methods for benchmarking studies.
\begin{itemize}
\item Greedy hypervolume subset selection (GHSS).  In our experiments, we use the lazy GHSS proposed in \cite{chen2021fast}\footnote{Source code available at \url{https://github.com/weiyuchen1999/LGISS}.}, which is the fastest implementation of GHSS.  The WFG algorithm \cite{While2012A} is used in GHSS for computing the hypervolume contribution of each solution.
\item Greedy approximated hypervolume subset selection (GAHSS) \cite{shang2021greedy}\footnote{Source code available at \url{https: //github.com/HisaoLabSUSTC/GAHSS}.}. This is an approximated version of GHSS by using hypervolume contribution approximation \cite{shang2018r2} instead of exact hypervolume contribution calculation in GHSS. 
\item Greedy IGD subset selection (GIGDSS). This is the greedy algorithm for IGDSS. We use the lazy GIGDSS proposed in \cite{chen2021fast} in our experiments since it is much faster than its non-lazy counterpart. 
\item Greedy IGD+ subset selection (GIGD+SS). This is the greedy algorithm for IGD+SS. We use the lazy GIGD+SS proposed in \cite{chen2021fast} in our experiments.
\item Distance-based subset selection (DSS). This is the original DSS method proposed in \cite{singh2018distance}, which is a greedy algorithm in nature.
\item Iterative distance-based subset selection (IDSS). This is an iterative DSS method proposed in \cite{shang2021distance}. The  maximum number of iterations (which is the termination condition) should be specified for IDSS.
\item Clustering-based subset selection using k-means (CSS-MEA). The use of k-means clustering for subset selection was  proposed in \cite{chen2021cluster}. The number of clusters is the same as the number of solutions to be selected (i.e., $k$). In general, each cluster center is not the same as any data point in k-means clustering. Thus, for each cluster, we choose a candidate solution which has the smallest total distance to all the other candidate solutions in the same cluster.
\item Clustering-based subset selection using k-medoids (CSS-MED). The use of k-medoids clustering for subset selection was  proposed in \cite{chen2021cluster}. Since each cluster center in k-medoids clustering is selected from the given data point, we can choose all the $k$ cluster centers as the result (i.e., selected subset) of subset selection.
\item Reference vector-based subset selection using the perpendicular distance (RVSS-PD). This selection method is the same as in NSGA-III. The Das and Dennis method is used to generate a set of reference vectors.
\item Reference vector-based subset selection using the angle distance (RVSS-AD). The only difference between RVSS-PD and RVSS-AD is that the angle distance is used in RVSS-AD.
\end{itemize}

These 10 methods belong to the first four types of subset selection methods as described in Section \ref{ss-methods}.
To be more specific, GHSS, GAHSS, GIGDSS and GIGD+SS are indicator-based subset selection methods, DSS and IDSS are distance-based subset selection methods, CSS-MEA and CSS-MED are clustering-based subset selection methods, and RVSS-PD and RVSS-AD are reference vector-based subset selection methods.

\subsubsection{Performance metrics}
In order to evaluate the performance of each subset selection method, we use the following performance metrics: hypervolume, IGD, IGD+, uniformity level\footnote{The uniformity level of a set is defined as the minimum distance between two points in this set \cite{sayin2000measuring}.}, and runtime since they are commonly used to evaluate the performance of EMO algorithms. For the hypervolume indicator, the reference point is set as 1.2 times the nadir point of the true Pareto front for the corresponding candidate solution set. For the IGD and IGD+ indicators, the reference point set is a set of uniformly distributed points on the true Pareto front of the corresponding candidate solution set\footnote{For the candidate solution sets on Pareto fronts, they are directly used as reference point sets for IGD and IGD+ evaluation. For the candidate solution sets from EMO algorithms, the reference point sets for IGD and IGD+ evaluation are generated using PlatEMO \cite{tian2017platemo}. The maximum number of reference points is set as 100K.}.

\subsubsection{Parameter settings}
\begin{itemize}
\item For the subset size $k$, we set $k$ as 91, 210, 156, and 275 for 3, 5, 8, and 10 objectives, respectively. This is because $k$ depends on the number of reference vectors in RVSS which cannot be set arbitrarily. We use the standard population size settings for $k$ in order to fairly compare all the methods. 
\item For the algorithms with randomness (i.e., IDSS, CSS-MEA, and CSS-MED), we perform 10 independent runs on each candidate solution set and report the average value. For the other algorithms, we perform a single run on each candidate solution set.
\item In GHSS and GAHSS, the reference point is set as 1.2 times the nadir point of the candidate solution set. In GAHSS, the number of direction vectors for the hypervolume contribution approximation is set as 100, as suggested in \cite{shang2021greedy}.
\item The maximum number of iterations  in IDSS, CSS-MEA, and CSS-MED is set as 1000. 
\item The maximum runtime for each algorithm on each candidate solution set is set as one hour. If the runtime of an algorithm exceeds one hour, we terminates this algorithm and no result is returned.
\end{itemize}

\subsubsection{Platforms}
We conduct the experiments on a PC equipped with ADM Ryzen Threadripper 3990X 64-Core CPU@2.90GHz, 256GB RAM and Windows 10 Operating System. All the subset selection methods are implemented in MATLAB R2021a.

\subsection{Results and Analysis}
\subsubsection{Candidate solution sets on Pareto fronts}
First, we run the 10 subset selection methods on the candidate solution sets on Pareto fronts. Tables \ref{tab:HV100000}-\ref{tab:time100000} show the results on the candidate solution sets with 100K solutions with respect to hypervolume, IGD, IGD+, uniformity level, and runtime, respectively. Results on the candidate solution sets with 10K and 1M solutions are provided in the supplementary material.

\begin{table*}[!htbp]
\fontsize{8.0pt}{0.4\baselineskip}\selectfont \setlength{\tabcolsep}{0.8pt}
\caption{The hypervolume performance of each subset selection method on each candidate solution set. The number of solutions in each candidate solution set is 100,000. The number in the parenthesis is the rank of the corresponding method among the 10 methods, where a smaller value indicates a better rank.}
\centering      \addtolength{\leftskip} {-2cm} \addtolength{\rightskip}{-2cm} 
       \begin{tabular}{cccccccccccc}
 \hline \multicolumn{2}{c}{\makecell{Candidate\\ Solution Set}}&\;\;\;\;GHSS\;\;\;\;&GAHSS&GIGDSS\;  &GIGD+SS&\;\;\;\;\;DSS\;\;\;\;\;&IDSS&CSS-MEA\;&CSS-MED\;&RVSS-PD\;&RVSS-AD\\ \hline 
\multirow{4}{*}{\makecell{Linear\\Triangular}}&3&\tiny{1.52E+0(3)}&\tiny{1.52E+0(4)}&\tiny{-(9.5)}&\tiny{-(9.5)}&\tiny{1.51E+0(5)}&\tiny{1.51E+0(6)}&\tiny{1.43E+0(8)}&\tiny{1.43E+0(7)}&\tiny{\textbf{1.52E+0(1)}}&\tiny{1.52E+0(2)}\\\cline{3-12} 
&5&\tiny{\textbf{2.46E+0(1)}}&\tiny{2.45E+0(2)}&\tiny{-(9.5)}&\tiny{-(9.5)}&\tiny{2.45E+0(3)}&\tiny{2.42E+0(6)}&\tiny{2.10E+0(8)}&\tiny{2.12E+0(7)}&\tiny{2.45E+0(5)}&\tiny{2.45E+0(4)}\\\cline{3-12} 
&8&\tiny{-(9)}&\tiny{\textbf{4.29E+0(1)}}&\tiny{-(9)}&\tiny{-(9)}&\tiny{4.28E+0(2)}&\tiny{4.21E+0(5)}&\tiny{3.56E+0(6)}&\tiny{3.53E+0(7)}&\tiny{4.27E+0(4)}&\tiny{4.28E+0(3)}\\\cline{3-12} 
&10&\tiny{-(9)}&\tiny{\textbf{6.19E+0(1)}}&\tiny{-(9)}&\tiny{-(9)}&\tiny{6.18E+0(2)}&\tiny{6.11E+0(5)}&\tiny{5.55E+0(6)}&\tiny{5.47E+0(7)}&\tiny{6.17E+0(4)}&\tiny{6.17E+0(3)}\\\hline 
\multirow{4}{*}{\makecell{Linear\\Inverted\\ Triangular}}&3&\tiny{\textbf{5.31E-1(1)}}&\tiny{5.29E-1(2)}&\tiny{-(9.5)}&\tiny{-(9.5)}&\tiny{5.25E-1(3)}&\tiny{5.22E-1(4)}&\tiny{5.11E-1(6)}&\tiny{5.11E-1(5)}&\tiny{5.03E-1(7)}&\tiny{5.03E-1(8)}\\\cline{3-12} 
&5&\tiny{\textbf{9.10E-2(1)}}&\tiny{9.05E-2(2)}&\tiny{-(9.5)}&\tiny{-(9.5)}&\tiny{8.97E-2(3)}&\tiny{8.49E-2(4)}&\tiny{7.99E-2(6)}&\tiny{8.02E-2(5)}&\tiny{7.78E-2(7)}&\tiny{6.96E-2(8)}\\\cline{3-12} 
&8&\tiny{\textbf{1.92E-3(1)}}&\tiny{1.87E-3(3)}&\tiny{-(9.5)}&\tiny{-(9.5)}&\tiny{1.88E-3(2)}&\tiny{1.82E-3(4)}&\tiny{1.58E-3(6)}&\tiny{1.59E-3(5)}&\tiny{1.06E-3(8)}&\tiny{1.21E-3(7)}\\\cline{3-12} 
&10&\tiny{-(9)}&\tiny{1.46E-4(2)}&\tiny{-(9)}&\tiny{-(9)}&\tiny{\textbf{1.48E-4(1)}}&\tiny{1.38E-4(3)}&\tiny{1.16E-4(4)}&\tiny{1.16E-4(5)}&\tiny{7.27E-5(7)}&\tiny{8.95E-5(6)}\\\hline 
\multirow{4}{*}{\makecell{Concave\\Triangular}}&3&\tiny{\textbf{1.15E+0(1)}}&\tiny{1.15E+0(2)}&\tiny{-(9.5)}&\tiny{-(9.5)}&\tiny{1.14E+0(3)}&\tiny{1.12E+0(6)}&\tiny{1.04E+0(8)}&\tiny{1.04E+0(7)}&\tiny{1.14E+0(4.5)}&\tiny{1.14E+0(4.5)}\\\cline{3-12} 
&5&\tiny{\textbf{2.20E+0(1)}}&\tiny{2.18E+0(2)}&\tiny{-(9.5)}&\tiny{-(9.5)}&\tiny{2.15E+0(5)}&\tiny{2.02E+0(6)}&\tiny{1.58E+0(7)}&\tiny{1.58E+0(8)}&\tiny{2.17E+0(3.5)}&\tiny{2.17E+0(3.5)}\\\cline{3-12} 
&8&\tiny{\textbf{4.09E+0(1)}}&\tiny{4.07E+0(2)}&\tiny{-(9.5)}&\tiny{-(9.5)}&\tiny{4.00E+0(5)}&\tiny{3.40E+0(6)}&\tiny{2.07E+0(7)}&\tiny{2.02E+0(8)}&\tiny{4.03E+0(3.5)}&\tiny{4.03E+0(3.5)}\\\cline{3-12} 
&10&\tiny{-(9)}&\tiny{\textbf{6.01E+0(1)}}&\tiny{-(9)}&\tiny{-(9)}&\tiny{5.93E+0(4)}&\tiny{4.97E+0(5)}&\tiny{3.22E+0(6)}&\tiny{2.92E+0(7)}&\tiny{5.94E+0(2.5)}&\tiny{5.94E+0(2.5)}\\\hline 
\multirow{4}{*}{\makecell{Concave\\Inverted\\ Triangular}}&3&\tiny{\textbf{2.26E-1(1)}}&\tiny{2.24E-1(2)}&\tiny{-(9.5)}&\tiny{-(9.5)}&\tiny{2.20E-1(3)}&\tiny{2.10E-1(6)}&\tiny{1.94E-1(8)}&\tiny{1.95E-1(7)}&\tiny{2.12E-1(5)}&\tiny{2.15E-1(4)}\\\cline{3-12} 
&5&\tiny{\textbf{2.02E-2(1)}}&\tiny{1.99E-2(2)}&\tiny{-(9.5)}&\tiny{-(9.5)}&\tiny{1.86E-2(3)}&\tiny{1.39E-2(4)}&\tiny{1.20E-2(6)}&\tiny{1.22E-2(5)}&\tiny{9.27E-3(7)}&\tiny{9.26E-3(8)}\\\cline{3-12} 
&8&\tiny{\textbf{2.34E-4(1)}}&\tiny{2.29E-4(2)}&\tiny{-(9.5)}&\tiny{-(9.5)}&\tiny{2.17E-4(3)}&\tiny{1.10E-4(5)}&\tiny{9.22E-5(7)}&\tiny{9.41E-5(6)}&\tiny{8.33E-5(8)}&\tiny{1.12E-4(4)}\\\cline{3-12} 
&10&\tiny{-(9)}&\tiny{\textbf{1.09E-5(1)}}&\tiny{-(9)}&\tiny{-(9)}&\tiny{1.02E-5(2)}&\tiny{4.53E-6(5)}&\tiny{5.17E-6(4)}&\tiny{5.96E-6(3)}&\tiny{2.82E-6(7)}&\tiny{3.17E-6(6)}\\\hline 
\multirow{4}{*}{\makecell{Convex\\Triangular}}&3&\tiny{\textbf{1.71E+0(1)}}&\tiny{1.71E+0(4)}&\tiny{-(9.5)}&\tiny{-(9.5)}&\tiny{1.71E+0(3)}&\tiny{1.71E+0(2)}&\tiny{1.71E+0(8)}&\tiny{1.71E+0(7)}&\tiny{1.71E+0(5)}&\tiny{1.71E+0(6)}\\\cline{3-12} 
&5&\tiny{\textbf{2.49E+0(1)}}&\tiny{2.49E+0(6)}&\tiny{-(9.5)}&\tiny{-(9.5)}&\tiny{2.49E+0(2)}&\tiny{2.49E+0(3)}&\tiny{2.49E+0(7)}&\tiny{2.49E+0(8)}&\tiny{2.49E+0(5)}&\tiny{2.49E+0(4)}\\\cline{3-12} 
&8&\tiny{-(9)}&\tiny{4.30E+0(5)}&\tiny{-(9)}&\tiny{-(9)}&\tiny{\textbf{4.30E+0(1)}}&\tiny{4.30E+0(3)}&\tiny{4.30E+0(6)}&\tiny{4.30E+0(7)}&\tiny{4.30E+0(4)}&\tiny{4.30E+0(2)}\\\cline{3-12} 
&10&\tiny{-(9)}&\tiny{6.19E+0(5)}&\tiny{-(9)}&\tiny{-(9)}&\tiny{\textbf{6.19E+0(1)}}&\tiny{6.19E+0(4)}&\tiny{6.19E+0(7)}&\tiny{6.19E+0(6)}&\tiny{6.19E+0(3)}&\tiny{6.19E+0(2)}\\\hline 
\multirow{4}{*}{\makecell{Convex\\Inverted\\ Triangular}}&3&\tiny{\textbf{1.04E+0(1)}}&\tiny{1.04E+0(4)}&\tiny{-(9.5)}&\tiny{-(9.5)}&\tiny{1.04E+0(6)}&\tiny{1.04E+0(5)}&\tiny{1.04E+0(2)}&\tiny{1.04E+0(3)}&\tiny{1.01E+0(8)}&\tiny{1.01E+0(7)}\\\cline{3-12} 
&5&\tiny{\textbf{5.01E-1(1)}}&\tiny{4.91E-1(2)}&\tiny{-(9.5)}&\tiny{-(9.5)}&\tiny{4.63E-1(6)}&\tiny{4.72E-1(5)}&\tiny{4.88E-1(4)}&\tiny{4.89E-1(3)}&\tiny{3.71E-1(7)}&\tiny{3.71E-1(8)}\\\cline{3-12} 
&8&\tiny{\textbf{6.43E-2(1)}}&\tiny{6.14E-2(2)}&\tiny{-(9.5)}&\tiny{-(9.5)}&\tiny{3.79E-2(6)}&\tiny{4.45E-2(5)}&\tiny{5.60E-2(4)}&\tiny{5.74E-2(3)}&\tiny{3.49E-2(7)}&\tiny{3.33E-2(8)}\\\cline{3-12} 
&10&\tiny{\textbf{1.47E-2(1)}}&\tiny{1.41E-2(2)}&\tiny{-(9.5)}&\tiny{-(9.5)}&\tiny{6.95E-3(8)}&\tiny{9.10E-3(5)}&\tiny{1.23E-2(4)}&\tiny{1.28E-2(3)}&\tiny{8.25E-3(6)}&\tiny{7.73E-3(7)}\\\hline 
\multicolumn{2}{c}{Avg Rank}&$3.42$&$\mathbf{2.54}$&$9.35$&$9.35$&$3.42$&$4.67$&$6.04$&$5.79$&$5.38$&$5.04$\\ \hline 
\end{tabular}
 \label{tab:HV100000}\end{table*}
 
 For the hypervolume performance in Table \ref{tab:HV100000}, we can see that GAHSS achieves the best average rank  followed by GHSS and DSS. However, GHSS has the best rank in almost all cases except for the cases where GHSS cannot obtain the results within one hour (and thus GHSS has the worst rank). GHSS and GAHSS show  the best performance because they aim to maximize the hypervolume of the subset. Since GAHSS is an approximated version of GHSS,GAHSS shows a slightly worse hypervolume performance than GHSS when the execution of GHSS is completed within one hour time limit. For GIGDSS and GIGD+SS, all results are not obtained within one hour, which means that these two methods are not applicable for large-scale subset selection. DSS and IDSS achieve medium hypervolume performance which is worse than GHSS and GAHSS, but better than all the other methods.

For the IGD performance in Table \ref{tab:IGD100000}, we can see that the best performance is achieved by CSS-MEA, followed by CSS-MED. This shows that the clustering-based subset selection is closely related to IGD-based subset selection. GIGDSS has a bad performance since all results are not obtained. This is because all the given candidate solutions are used as reference points for IGD calculation. When the candidate solution sets are small (i.e., 10K in the supplementary file instead of 100K in this paper), better IGD performance is obtained by the IGD-based method (GIGDSS), which is similar to the clustering-based methods (CSS-MEA and CSS-MED). Table \ref{tab:IGD100000} suggests that the clustering-based subset selection methods are more suitable for large-scale IGD-based subset selection.  Table \ref{tab:IGD100000} also suggests that GIGDSS needs some modification to decrease its computation time (i.e., the use of less reference points for IGD calculation). DSS and IDSS achieve medium IGD performance among the 10 methods.
\begin{table*}[!tbp]
\fontsize{8.0pt}{0.4\baselineskip}\selectfont
\setlength{\tabcolsep}{0.8pt}
\caption{The IGD performance of each subset selection method on each candidate solution set. The number of solutions in each candidate solution set is 100,000. The number in the parenthesis is the rank of the corresponding method among the 10 methods, where a smaller value indicates a better rank.}
\centering      \addtolength{\leftskip} {-2cm} \addtolength{\rightskip}{-2cm} 
       \begin{tabular}{cccccccccccc}
 \hline \multicolumn{2}{c}{\makecell{Candidate\\ Solution Set}}&\;\;\;\;GHSS\;\;\;\;&GAHSS&GIGDSS\;&GIGD+SS&\;\;\;\;\;DSS\;\;\;\;\;&IDSS&CSS-MEA\;&CSS-MED\;&RVSS-PD\;&RVSS-AD\\ \hline 
\multirow{4}{*}{\makecell{Linear\\Triangular}}&3&\tiny{4.27E-2(7)}&\tiny{4.24E-2(6)}&\tiny{-(9.5)}&\tiny{-(9.5)}&\tiny{4.38E-2(8)}&\tiny{4.04E-2(3)}&\tiny{\textbf{3.70E-2(1)}}&\tiny{3.71E-2(2)}&\tiny{4.12E-2(4)}&\tiny{4.12E-2(5)}\\\cline{3-12} 
&5&\tiny{1.06E-1(8)}&\tiny{1.01E-1(5)}&\tiny{-(9.5)}&\tiny{-(9.5)}&\tiny{9.86E-2(4)}&\tiny{9.25E-2(3)}&\tiny{\textbf{8.17E-2(1)}}&\tiny{8.20E-2(2)}&\tiny{1.04E-1(6)}&\tiny{1.04E-1(7)}\\\cline{3-12} 
&8&\tiny{-(9)}&\tiny{1.85E-1(7)}&\tiny{-(9)}&\tiny{-(9)}&\tiny{1.77E-1(6)}&\tiny{1.57E-1(3)}&\tiny{1.33E-1(2)}&\tiny{\textbf{1.32E-1(1)}}&\tiny{1.72E-1(4)}&\tiny{1.74E-1(5)}\\\cline{3-12} 
&10&\tiny{-(9)}&\tiny{1.88E-1(7)}&\tiny{-(9)}&\tiny{-(9)}&\tiny{1.76E-1(4)}&\tiny{1.57E-1(3)}&\tiny{1.36E-1(2)}&\tiny{\textbf{1.36E-1(1)}}&\tiny{1.77E-1(5)}&\tiny{1.83E-1(6)}\\\hline 
\multirow{4}{*}{\makecell{Linear\\Inverted\\ Triangular}}&3&\tiny{4.53E-2(6)}&\tiny{4.37E-2(4)}&\tiny{-(9.5)}&\tiny{-(9.5)}&\tiny{4.39E-2(5)}&\tiny{4.03E-2(3)}&\tiny{\textbf{3.71E-2(1)}}&\tiny{3.72E-2(2)}&\tiny{7.07E-2(7)}&\tiny{7.10E-2(8)}\\\cline{3-12} 
&5&\tiny{1.05E-1(6)}&\tiny{1.05E-1(5)}&\tiny{-(9.5)}&\tiny{-(9.5)}&\tiny{9.89E-2(4)}&\tiny{9.25E-2(3)}&\tiny{\textbf{8.16E-2(1)}}&\tiny{8.19E-2(2)}&\tiny{1.32E-1(7)}&\tiny{1.63E-1(8)}\\\cline{3-12} 
&8&\tiny{1.67E-1(5)}&\tiny{1.60E-1(4)}&\tiny{-(9.5)}&\tiny{-(9.5)}&\tiny{1.76E-1(6)}&\tiny{1.57E-1(3)}&\tiny{1.33E-1(2)}&\tiny{\textbf{1.32E-1(1)}}&\tiny{1.99E-1(8)}&\tiny{1.91E-1(7)}\\\cline{3-12} 
&10&\tiny{-(9)}&\tiny{1.63E-1(4)}&\tiny{-(9)}&\tiny{-(9)}&\tiny{1.74E-1(5)}&\tiny{1.57E-1(3)}&\tiny{1.36E-1(2)}&\tiny{\textbf{1.35E-1(1)}}&\tiny{1.87E-1(7)}&\tiny{1.81E-1(6)}\\\hline 
\multirow{4}{*}{\makecell{Concave\\Triangular}}&3&\tiny{7.45E-2(8)}&\tiny{6.38E-2(7)}&\tiny{-(9.5)}&\tiny{-(9.5)}&\tiny{5.53E-2(4)}&\tiny{5.36E-2(3)}&\tiny{\textbf{4.99E-2(1)}}&\tiny{5.00E-2(2)}&\tiny{5.78E-2(5.5)}&\tiny{5.78E-2(5.5)}\\\cline{3-12} 
&5&\tiny{2.24E-1(8)}&\tiny{1.83E-1(5)}&\tiny{-(9.5)}&\tiny{-(9.5)}&\tiny{1.58E-1(4)}&\tiny{1.54E-1(3)}&\tiny{\textbf{1.40E-1(1)}}&\tiny{1.40E-1(2)}&\tiny{1.83E-1(6.5)}&\tiny{1.83E-1(6.5)}\\\cline{3-12} 
&8&\tiny{4.19E-1(8)}&\tiny{3.88E-1(7)}&\tiny{-(9.5)}&\tiny{-(9.5)}&\tiny{3.30E-1(4)}&\tiny{3.13E-1(3)}&\tiny{2.78E-1(2)}&\tiny{\textbf{2.77E-1(1)}}&\tiny{3.78E-1(5.5)}&\tiny{3.78E-1(5.5)}\\\cline{3-12} 
&10&\tiny{-(9)}&\tiny{4.25E-1(7)}&\tiny{-(9)}&\tiny{-(9)}&\tiny{3.64E-1(4)}&\tiny{3.47E-1(3)}&\tiny{3.14E-1(2)}&\tiny{\textbf{3.12E-1(1)}}&\tiny{4.18E-1(5.5)}&\tiny{4.18E-1(5.5)}\\\hline 
\multirow{4}{*}{\makecell{Concave\\Inverted\\ Triangular}}&3&\tiny{3.68E-2(6)}&\tiny{3.50E-2(5)}&\tiny{-(9.5)}&\tiny{-(9.5)}&\tiny{2.83E-2(4)}&\tiny{2.69E-2(3)}&\tiny{\textbf{2.43E-2(1)}}&\tiny{2.43E-2(2)}&\tiny{6.86E-2(8)}&\tiny{6.64E-2(7)}\\\cline{3-12} 
&5&\tiny{6.48E-2(6)}&\tiny{5.62E-2(5)}&\tiny{-(9.5)}&\tiny{-(9.5)}&\tiny{4.32E-2(4)}&\tiny{3.74E-2(3)}&\tiny{\textbf{3.20E-2(1)}}&\tiny{3.22E-2(2)}&\tiny{2.75E-1(8)}&\tiny{1.79E-1(7)}\\\cline{3-12} 
&8&\tiny{9.76E-2(6)}&\tiny{8.50E-2(5)}&\tiny{-(9.5)}&\tiny{-(9.5)}&\tiny{6.43E-2(4)}&\tiny{4.69E-2(3)}&\tiny{\textbf{3.68E-2(1)}}&\tiny{3.69E-2(2)}&\tiny{3.20E-1(8)}&\tiny{2.46E-1(7)}\\\cline{3-12} 
&10&\tiny{-(9)}&\tiny{7.22E-2(5)}&\tiny{-(9)}&\tiny{-(9)}&\tiny{5.35E-2(4)}&\tiny{3.86E-2(3)}&\tiny{\textbf{3.11E-2(1)}}&\tiny{3.11E-2(2)}&\tiny{5.19E-1(7)}&\tiny{2.20E-1(6)}\\\hline 
\multirow{4}{*}{\makecell{Convex\\Triangular}}&3&\tiny{2.78E-2(4)}&\tiny{2.94E-2(6)}&\tiny{-(9.5)}&\tiny{-(9.5)}&\tiny{2.85E-2(5)}&\tiny{2.69E-2(3)}&\tiny{\textbf{2.42E-2(1)}}&\tiny{2.43E-2(2)}&\tiny{4.46E-2(7)}&\tiny{4.46E-2(8)}\\\cline{3-12} 
&5&\tiny{3.79E-2(5)}&\tiny{3.60E-2(3)}&\tiny{-(9.5)}&\tiny{-(9.5)}&\tiny{4.32E-2(6)}&\tiny{3.73E-2(4)}&\tiny{\textbf{3.20E-2(1)}}&\tiny{3.21E-2(2)}&\tiny{5.09E-2(7)}&\tiny{5.11E-2(8)}\\\cline{3-12} 
&8&\tiny{-(9)}&\tiny{4.53E-2(3)}&\tiny{-(9)}&\tiny{-(9)}&\tiny{6.38E-2(7)}&\tiny{4.68E-2(4)}&\tiny{\textbf{3.69E-2(1)}}&\tiny{3.69E-2(2)}&\tiny{5.40E-2(6)}&\tiny{5.35E-2(5)}\\\cline{3-12} 
&10&\tiny{-(9)}&\tiny{3.66E-2(3)}&\tiny{-(9)}&\tiny{-(9)}&\tiny{5.37E-2(7)}&\tiny{3.83E-2(4)}&\tiny{\textbf{3.10E-2(1)}}&\tiny{3.11E-2(2)}&\tiny{4.15E-2(5)}&\tiny{4.19E-2(6)}\\\hline 
\multirow{4}{*}{\makecell{Convex\\Inverted\\ Triangular}}&3&\tiny{5.45E-2(5)}&\tiny{5.44E-2(4)}&\tiny{-(9.5)}&\tiny{-(9.5)}&\tiny{5.46E-2(6)}&\tiny{5.38E-2(3)}&\tiny{\textbf{4.99E-2(1)}}&\tiny{5.00E-2(2)}&\tiny{7.46E-2(8)}&\tiny{7.46E-2(7)}\\\cline{3-12} 
&5&\tiny{1.54E-1(5)}&\tiny{1.49E-1(3)}&\tiny{-(9.5)}&\tiny{-(9.5)}&\tiny{1.59E-1(6)}&\tiny{1.54E-1(4)}&\tiny{\textbf{1.40E-1(1)}}&\tiny{1.40E-1(2)}&\tiny{2.06E-1(7)}&\tiny{2.06E-1(8)}\\\cline{3-12} 
&8&\tiny{2.99E-1(4)}&\tiny{2.89E-1(3)}&\tiny{-(9.5)}&\tiny{-(9.5)}&\tiny{3.28E-1(6)}&\tiny{3.13E-1(5)}&\tiny{2.78E-1(2)}&\tiny{\textbf{2.77E-1(1)}}&\tiny{3.79E-1(7)}&\tiny{3.81E-1(8)}\\\cline{3-12} 
&10&\tiny{3.31E-1(4)}&\tiny{3.26E-1(3)}&\tiny{-(9.5)}&\tiny{-(9.5)}&\tiny{3.68E-1(6)}&\tiny{3.47E-1(5)}&\tiny{3.15E-1(2)}&\tiny{\textbf{3.12E-1(1)}}&\tiny{4.03E-1(8)}&\tiny{4.00E-1(7)}\\\hline 
\multicolumn{2}{c}{Avg Rank}&$6.83$&$4.83$&$9.35$&$9.35$&$5.12$&$3.33$&$\mathbf{1.33}$&$1.67$&$6.54$&$6.62$\\ \hline 
\end{tabular}
 \label{tab:IGD100000}\end{table*}

For the IGD+ performance in Table \ref{tab:IGDp100000}, we can see that the best performance is achieved by GAHSS, followed by CSS-MED and CSS-MEA. If we take a closer look at the results, we can see that GHSS performs the best on some solution sets (e.g., concave triangular and convex inverted triangular). This is because IGD+ is closely related to hypervolume with respect to their optimal solution distributions \cite{ishibuchi2019comparison}. DSS and IDSS still have medial IGD+ performance among the 10 methods. GIGDSS, GIGD+SS also needs some modification to decrease its computation time for its application to large-scale subset selection. For small candidate solution sets in the supplementary file, good IGD+ performance is obtained by GIGD+SS. 
\begin{table*}[!tbp]
\fontsize{8.0pt}{0.4\baselineskip}\selectfont
\setlength{\tabcolsep}{0.8pt}
\caption{The IGD+ performance of each subset selection method on each candidate solution set. The number of solutions in each candidate solution set is 100,000. The number in the parenthesis is the rank of the corresponding method among the 10 methods, where a smaller value indicates a better rank.}
\centering      \addtolength{\leftskip} {-2cm} \addtolength{\rightskip}{-2cm} 
       \begin{tabular}{cccccccccccc}
 \hline \multicolumn{2}{c}{\makecell{Candidate\\ Solution Set}}&\;\;\;\;GHSS\;\;\;\;&GAHSS&GIGDSS\;&GIGD+SS&\;\;\;\;\;DSS\;\;\;\;\;&IDSS&CSS-MEA\;&CSS-MED\;&RVSS-PD\;&RVSS-AD\\ \hline 
\multirow{4}{*}{\makecell{Linear\\Triangular}}&3&\tiny{2.96E-2(7)}&\tiny{2.94E-2(6)}&\tiny{-(9.5)}&\tiny{-(9.5)}&\tiny{3.04E-2(8)}&\tiny{2.81E-2(3)}&\tiny{\textbf{2.57E-2(1)}}&\tiny{2.57E-2(2)}&\tiny{2.92E-2(4)}&\tiny{2.92E-2(5)}\\\cline{3-12} 
&5&\tiny{7.35E-2(6)}&\tiny{7.02E-2(5)}&\tiny{-(9.5)}&\tiny{-(9.5)}&\tiny{6.89E-2(4)}&\tiny{6.33E-2(3)}&\tiny{5.51E-2(2)}&\tiny{\textbf{5.50E-2(1)}}&\tiny{7.50E-2(7)}&\tiny{7.52E-2(8)}\\\cline{3-12} 
&8&\tiny{-(9)}&\tiny{1.31E-1(7)}&\tiny{-(9)}&\tiny{-(9)}&\tiny{1.25E-1(6)}&\tiny{1.08E-1(3)}&\tiny{8.60E-2(2)}&\tiny{\textbf{8.56E-2(1)}}&\tiny{1.20E-1(4)}&\tiny{1.22E-1(5)}\\\cline{3-12} 
&10&\tiny{-(9)}&\tiny{1.34E-1(7)}&\tiny{-(9)}&\tiny{-(9)}&\tiny{1.24E-1(4)}&\tiny{1.07E-1(3)}&\tiny{8.74E-2(2)}&\tiny{\textbf{8.66E-2(1)}}&\tiny{1.25E-1(5)}&\tiny{1.32E-1(6)}\\\hline 
\multirow{4}{*}{\makecell{Linear\\Inverted\\ Triangular}}&3&\tiny{3.12E-2(6)}&\tiny{3.01E-2(5)}&\tiny{-(9.5)}&\tiny{-(9.5)}&\tiny{2.98E-2(4)}&\tiny{2.79E-2(3)}&\tiny{\textbf{2.61E-2(1)}}&\tiny{2.61E-2(2)}&\tiny{5.15E-2(7)}&\tiny{5.15E-2(8)}\\\cline{3-12} 
&5&\tiny{6.98E-2(5)}&\tiny{7.00E-2(6)}&\tiny{-(9.5)}&\tiny{-(9.5)}&\tiny{6.61E-2(4)}&\tiny{6.27E-2(3)}&\tiny{5.84E-2(2)}&\tiny{\textbf{5.84E-2(1)}}&\tiny{8.49E-2(7)}&\tiny{1.03E-1(8)}\\\cline{3-12} 
&8&\tiny{1.09E-1(5)}&\tiny{1.08E-1(4)}&\tiny{-(9.5)}&\tiny{-(9.5)}&\tiny{1.14E-1(6)}&\tiny{1.05E-1(3)}&\tiny{9.70E-2(2)}&\tiny{\textbf{9.68E-2(1)}}&\tiny{1.34E-1(8)}&\tiny{1.26E-1(7)}\\\cline{3-12} 
&10&\tiny{-(9)}&\tiny{1.09E-1(4)}&\tiny{-(9)}&\tiny{-(9)}&\tiny{1.13E-1(5)}&\tiny{1.05E-1(3)}&\tiny{9.90E-2(2)}&\tiny{\textbf{9.88E-2(1)}}&\tiny{1.30E-1(7)}&\tiny{1.21E-1(6)}\\\hline 
\multirow{4}{*}{\makecell{Concave\\Triangular}}&3&\tiny{\textbf{2.26E-2(1)}}&\tiny{2.46E-2(2)}&\tiny{-(9.5)}&\tiny{-(9.5)}&\tiny{2.54E-2(3)}&\tiny{2.61E-2(4)}&\tiny{2.73E-2(7)}&\tiny{2.75E-2(8)}&\tiny{2.61E-2(5.5)}&\tiny{2.61E-2(5.5)}\\\cline{3-12} 
&5&\tiny{\textbf{7.12E-2(1)}}&\tiny{7.42E-2(2)}&\tiny{-(9.5)}&\tiny{-(9.5)}&\tiny{7.79E-2(3)}&\tiny{8.27E-2(6)}&\tiny{8.76E-2(8)}&\tiny{8.76E-2(7)}&\tiny{7.86E-2(4.5)}&\tiny{7.86E-2(4.5)}\\\cline{3-12} 
&8&\tiny{1.70E-1(2)}&\tiny{\textbf{1.70E-1(1)}}&\tiny{-(9.5)}&\tiny{-(9.5)}&\tiny{1.73E-1(5)}&\tiny{1.82E-1(6)}&\tiny{1.93E-1(8)}&\tiny{1.93E-1(7)}&\tiny{1.70E-1(3.5)}&\tiny{1.70E-1(3.5)}\\\cline{3-12} 
&10&\tiny{-(9)}&\tiny{\textbf{1.96E-1(1)}}&\tiny{-(9)}&\tiny{-(9)}&\tiny{2.00E-1(4)}&\tiny{2.09E-1(5)}&\tiny{2.18E-1(6)}&\tiny{2.19E-1(7)}&\tiny{1.96E-1(2.5)}&\tiny{1.96E-1(2.5)}\\\hline 
\multirow{4}{*}{\makecell{Concave\\Inverted\\ Triangular}}&3&\tiny{1.12E-2(4)}&\tiny{\textbf{9.92E-3(1)}}&\tiny{-(9.5)}&\tiny{-(9.5)}&\tiny{1.03E-2(2)}&\tiny{1.09E-2(3)}&\tiny{1.37E-2(6)}&\tiny{1.35E-2(5)}&\tiny{1.90E-2(8)}&\tiny{1.83E-2(7)}\\\cline{3-12} 
&5&\tiny{1.60E-2(2)}&\tiny{\textbf{1.51E-2(1)}}&\tiny{-(9.5)}&\tiny{-(9.5)}&\tiny{1.60E-2(3)}&\tiny{1.75E-2(4)}&\tiny{2.21E-2(6)}&\tiny{2.17E-2(5)}&\tiny{6.93E-2(8)}&\tiny{5.99E-2(7)}\\\cline{3-12} 
&8&\tiny{2.39E-2(4)}&\tiny{\textbf{2.17E-2(1)}}&\tiny{-(9.5)}&\tiny{-(9.5)}&\tiny{2.21E-2(2)}&\tiny{2.32E-2(3)}&\tiny{2.79E-2(6)}&\tiny{2.78E-2(5)}&\tiny{6.03E-2(8)}&\tiny{5.14E-2(7)}\\\cline{3-12} 
&10&\tiny{-(9)}&\tiny{1.92E-2(2)}&\tiny{-(9)}&\tiny{-(9)}&\tiny{\textbf{1.89E-2(1)}}&\tiny{2.04E-2(3)}&\tiny{2.36E-2(4)}&\tiny{2.37E-2(5)}&\tiny{6.72E-2(7)}&\tiny{5.48E-2(6)}\\\hline 
\multirow{4}{*}{\makecell{Convex\\Triangular}}&3&\tiny{1.15E-2(4)}&\tiny{\textbf{1.10E-2(1)}}&\tiny{-(9.5)}&\tiny{-(9.5)}&\tiny{1.52E-2(6)}&\tiny{1.35E-2(5)}&\tiny{1.11E-2(2)}&\tiny{1.12E-2(3)}&\tiny{2.11E-2(7)}&\tiny{2.11E-2(8)}\\\cline{3-12} 
&5&\tiny{1.88E-2(4)}&\tiny{1.57E-2(3)}&\tiny{-(9.5)}&\tiny{-(9.5)}&\tiny{2.74E-2(6)}&\tiny{2.11E-2(5)}&\tiny{\textbf{1.37E-2(1)}}&\tiny{1.39E-2(2)}&\tiny{3.04E-2(7)}&\tiny{3.12E-2(8)}\\\cline{3-12} 
&8&\tiny{-(9)}&\tiny{2.26E-2(5)}&\tiny{-(9)}&\tiny{-(9)}&\tiny{2.87E-2(6)}&\tiny{3.08E-2(7)}&\tiny{1.50E-2(2)}&\tiny{\textbf{1.49E-2(1)}}&\tiny{1.68E-2(3)}&\tiny{1.70E-2(4)}\\\cline{3-12} 
&10&\tiny{-(9)}&\tiny{1.93E-2(5)}&\tiny{-(9)}&\tiny{-(9)}&\tiny{2.35E-2(6)}&\tiny{2.40E-2(7)}&\tiny{\textbf{1.27E-2(1)}}&\tiny{1.27E-2(2)}&\tiny{1.37E-2(3)}&\tiny{1.49E-2(4)}\\\hline 
\multirow{4}{*}{\makecell{Convex\\Inverted\\ Triangular}}&3&\tiny{\textbf{2.59E-2(1)}}&\tiny{2.68E-2(4)}&\tiny{-(9.5)}&\tiny{-(9.5)}&\tiny{2.95E-2(6)}&\tiny{2.90E-2(5)}&\tiny{2.64E-2(2)}&\tiny{2.65E-2(3)}&\tiny{4.30E-2(8)}&\tiny{4.30E-2(7)}\\\cline{3-12} 
&5&\tiny{\textbf{7.30E-2(1)}}&\tiny{7.88E-2(2)}&\tiny{-(9.5)}&\tiny{-(9.5)}&\tiny{9.65E-2(6)}&\tiny{9.10E-2(5)}&\tiny{8.04E-2(3)}&\tiny{8.05E-2(4)}&\tiny{1.46E-1(7)}&\tiny{1.47E-1(8)}\\\cline{3-12} 
&8&\tiny{\textbf{1.49E-1(1)}}&\tiny{1.56E-1(2)}&\tiny{-(9.5)}&\tiny{-(9.5)}&\tiny{2.22E-1(6)}&\tiny{2.05E-1(5)}&\tiny{1.70E-1(4)}&\tiny{1.67E-1(3)}&\tiny{2.27E-1(7)}&\tiny{2.37E-1(8)}\\\cline{3-12} 
&10&\tiny{\textbf{1.69E-1(1)}}&\tiny{1.74E-1(2)}&\tiny{-(9.5)}&\tiny{-(9.5)}&\tiny{2.51E-1(8)}&\tiny{2.28E-1(5)}&\tiny{1.91E-1(4)}&\tiny{1.88E-1(3)}&\tiny{2.29E-1(6)}&\tiny{2.41E-1(7)}\\\hline 
\multicolumn{2}{c}{Avg Rank}&$4.92$&$\mathbf{3.29}$&$9.35$&$9.35$&$4.75$&$4.25$&$3.50$&$3.33$&$6.00$&$6.25$\\ \hline 
\end{tabular}
 \label{tab:IGDp100000}\end{table*}

For the uniformity level performance in Table \ref{tab:uniformLevel100000}, DSS performs the best, followed by IDSS. This is because these two methods aim to maximize the uniformity level of the subset. CSS-MEA and CSS-MED have medial uniformity level performance. This is because the clustering-based methods are closely related to IGD minimization. However, the optimal distribution of solutions for IGD minimization does not include any solution on the boundary of the Pareto front (i.e., it is inside the Pareto front \cite{ishibuchi2018reference,ishibuchi2019comparison}). Thus, the uniformity level of the selected subsets by the clustering-based methods can be further improved by pushing some solutions towards the boundary of the Pareto front. 
 \begin{table*}[!tbp]
\fontsize{8.0pt}{0.4\baselineskip}\selectfont
\setlength{\tabcolsep}{0.8pt}
\caption{The uniformity level performance of each subset selection method on each candidate solution set. The number of solutions in each candidate solution set is 100,000. The number in the parenthesis is the rank of the corresponding method among the 10 methods, where a smaller value indicates a better rank.}
\centering      \addtolength{\leftskip} {-2cm} \addtolength{\rightskip}{-2cm} 
       \begin{tabular}{cccccccccccc}
 \hline \multicolumn{2}{c}{\makecell{Candidate\\ Solution Set}}&\;\;\;\;GHSS\;\;\;\;&GAHSS&GIGDSS\;&GIGD+SS&\;\;\;\;\;DSS\;\;\;\;\;&IDSS&CSS-MEA\;&CSS-MED\;&RVSS-PD\;&RVSS-AD\\ \hline 
\multirow{4}{*}{\makecell{Linear\\Triangular}}&3&\tiny{4.29E-2(8)}&\tiny{5.04E-2(7)}&\tiny{-(9.5)}&\tiny{-(9.5)}&\tiny{8.92E-2(3)}&\tiny{8.10E-2(5)}&\tiny{8.15E-2(4)}&\tiny{7.84E-2(6)}&\tiny{\textbf{1.11E-1(1)}}&\tiny{1.10E-1(2)}\\\cline{3-12} 
&5&\tiny{1.04E-1(7)}&\tiny{1.19E-1(5)}&\tiny{-(9.5)}&\tiny{-(9.5)}&\tiny{1.62E-1(3)}&\tiny{1.35E-1(4)}&\tiny{1.02E-1(8)}&\tiny{1.08E-1(6)}&\tiny{\textbf{1.62E-1(1.5)}}&\tiny{\textbf{1.62E-1(1.5)}}\\\cline{3-12} 
&8&\tiny{-(9)}&\tiny{1.43E-1(5)}&\tiny{-(9)}&\tiny{-(9)}&\tiny{\textbf{2.58E-1(1)}}&\tiny{2.07E-1(2)}&\tiny{1.07E-1(7)}&\tiny{1.18E-1(6)}&\tiny{1.51E-1(3.5)}&\tiny{1.51E-1(3.5)}\\\cline{3-12} 
&10&\tiny{-(9)}&\tiny{1.43E-1(5)}&\tiny{-(9)}&\tiny{-(9)}&\tiny{\textbf{2.46E-1(1)}}&\tiny{1.92E-1(2)}&\tiny{8.91E-2(7)}&\tiny{1.02E-1(6)}&\tiny{1.51E-1(4)}&\tiny{1.58E-1(3)}\\\hline 
\multirow{4}{*}{\makecell{Linear\\Inverted\\ Triangular}}&3&\tiny{4.98E-2(6)}&\tiny{5.76E-2(5)}&\tiny{-(9.5)}&\tiny{-(9.5)}&\tiny{\textbf{8.93E-2(1)}}&\tiny{8.09E-2(2)}&\tiny{8.00E-2(3)}&\tiny{7.91E-2(4)}&\tiny{0(7.5)}&\tiny{0(7.5)}\\\cline{3-12} 
&5&\tiny{9.49E-2(6)}&\tiny{1.02E-1(4)}&\tiny{-(9.5)}&\tiny{-(9.5)}&\tiny{\textbf{1.64E-1(1)}}&\tiny{1.35E-1(2)}&\tiny{1.00E-1(5)}&\tiny{1.05E-1(3)}&\tiny{0(7.5)}&\tiny{0(7.5)}\\\cline{3-12} 
&8&\tiny{1.63E-1(4)}&\tiny{1.85E-1(3)}&\tiny{-(9.5)}&\tiny{-(9.5)}&\tiny{\textbf{2.60E-1(1)}}&\tiny{2.07E-1(2)}&\tiny{1.14E-1(6)}&\tiny{1.16E-1(5)}&\tiny{0(7.5)}&\tiny{0(7.5)}\\\cline{3-12} 
&10&\tiny{-(9)}&\tiny{1.50E-1(3)}&\tiny{-(9)}&\tiny{-(9)}&\tiny{\textbf{2.46E-1(1)}}&\tiny{1.93E-1(2)}&\tiny{8.97E-2(5)}&\tiny{1.02E-1(4)}&\tiny{0(6.5)}&\tiny{0(6.5)}\\\hline 
\multirow{4}{*}{\makecell{Concave\\Triangular}}&3&\tiny{5.09E-2(8)}&\tiny{6.73E-2(7)}&\tiny{-(9.5)}&\tiny{-(9.5)}&\tiny{\textbf{1.14E-1(1)}}&\tiny{1.06E-1(3)}&\tiny{1.07E-1(2)}&\tiny{1.05E-1(4)}&\tiny{8.89E-2(5.5)}&\tiny{8.89E-2(5.5)}\\\cline{3-12} 
&5&\tiny{6.76E-2(8)}&\tiny{1.30E-1(7)}&\tiny{-(9.5)}&\tiny{-(9.5)}&\tiny{\textbf{2.62E-1(1)}}&\tiny{2.23E-1(2)}&\tiny{1.78E-1(4)}&\tiny{1.83E-1(3)}&\tiny{1.58E-1(5.5)}&\tiny{1.58E-1(5.5)}\\\cline{3-12} 
&8&\tiny{2.08E-1(8)}&\tiny{2.40E-1(4)}&\tiny{-(9.5)}&\tiny{-(9.5)}&\tiny{\textbf{4.81E-1(1)}}&\tiny{4.07E-1(2)}&\tiny{2.26E-1(5)}&\tiny{2.67E-1(3)}&\tiny{2.12E-1(6.5)}&\tiny{2.12E-1(6.5)}\\\cline{3-12} 
&10&\tiny{-(9)}&\tiny{2.53E-1(4)}&\tiny{-(9)}&\tiny{-(9)}&\tiny{\textbf{4.98E-1(1)}}&\tiny{4.20E-1(2)}&\tiny{2.31E-1(5)}&\tiny{2.58E-1(3)}&\tiny{1.54E-1(6.5)}&\tiny{1.54E-1(6.5)}\\\hline 
\multirow{4}{*}{\makecell{Concave\\Inverted\\ Triangular}}&3&\tiny{1.17E-2(6)}&\tiny{3.02E-2(5)}&\tiny{-(9.5)}&\tiny{-(9.5)}&\tiny{\textbf{5.61E-2(1)}}&\tiny{5.50E-2(2)}&\tiny{4.79E-2(4)}&\tiny{4.85E-2(3)}&\tiny{0(7.5)}&\tiny{0(7.5)}\\\cline{3-12} 
&5&\tiny{2.38E-2(6)}&\tiny{2.83E-2(5)}&\tiny{-(9.5)}&\tiny{-(9.5)}&\tiny{\textbf{7.46E-2(1)}}&\tiny{5.68E-2(2)}&\tiny{4.36E-2(3)}&\tiny{4.26E-2(4)}&\tiny{0(7.5)}&\tiny{0(7.5)}\\\cline{3-12} 
&8&\tiny{2.84E-2(6)}&\tiny{6.42E-2(3)}&\tiny{-(9.5)}&\tiny{-(9.5)}&\tiny{\textbf{1.05E-1(1)}}&\tiny{6.53E-2(2)}&\tiny{3.57E-2(5)}&\tiny{3.65E-2(4)}&\tiny{0(7.5)}&\tiny{0(7.5)}\\\cline{3-12} 
&10&\tiny{-(9)}&\tiny{3.23E-2(3)}&\tiny{-(9)}&\tiny{-(9)}&\tiny{\textbf{8.18E-2(1)}}&\tiny{4.88E-2(2)}&\tiny{2.56E-2(5)}&\tiny{2.58E-2(4)}&\tiny{0(6.5)}&\tiny{0(6.5)}\\\hline 
\multirow{4}{*}{\makecell{Convex\\Triangular}}&3&\tiny{3.42E-2(7)}&\tiny{3.23E-2(8)}&\tiny{-(9.5)}&\tiny{-(9.5)}&\tiny{\textbf{5.42E-2(1)}}&\tiny{5.41E-2(2)}&\tiny{4.68E-2(4)}&\tiny{4.69E-2(3)}&\tiny{3.76E-2(5.5)}&\tiny{3.76E-2(5.5)}\\\cline{3-12} 
&5&\tiny{3.25E-2(8)}&\tiny{3.78E-2(5)}&\tiny{-(9.5)}&\tiny{-(9.5)}&\tiny{\textbf{7.40E-2(1)}}&\tiny{5.65E-2(2)}&\tiny{4.28E-2(3)}&\tiny{4.19E-2(4)}&\tiny{3.63E-2(6.5)}&\tiny{3.63E-2(6.5)}\\\cline{3-12} 
&8&\tiny{-(9)}&\tiny{4.18E-2(4)}&\tiny{-(9)}&\tiny{-(9)}&\tiny{\textbf{1.03E-1(1)}}&\tiny{6.50E-2(2)}&\tiny{3.62E-2(7)}&\tiny{3.75E-2(5)}&\tiny{3.69E-2(6)}&\tiny{4.71E-2(3)}\\\cline{3-12} 
&10&\tiny{-(9)}&\tiny{3.05E-2(4)}&\tiny{-(9)}&\tiny{-(9)}&\tiny{\textbf{8.25E-2(1)}}&\tiny{4.87E-2(2)}&\tiny{2.47E-2(6)}&\tiny{2.46E-2(7)}&\tiny{2.50E-2(5)}&\tiny{3.30E-2(3)}\\\hline 
\multirow{4}{*}{\makecell{Convex\\Inverted\\ Triangular}}&3&\tiny{7.05E-2(6)}&\tiny{7.50E-2(5)}&\tiny{-(9.5)}&\tiny{-(9.5)}&\tiny{\textbf{1.10E-1(1)}}&\tiny{1.07E-1(4)}&\tiny{1.10E-1(2)}&\tiny{1.08E-1(3)}&\tiny{0(7.5)}&\tiny{0(7.5)}\\\cline{3-12} 
&5&\tiny{1.40E-1(5)}&\tiny{1.32E-1(6)}&\tiny{-(9.5)}&\tiny{-(9.5)}&\tiny{\textbf{2.64E-1(1)}}&\tiny{2.23E-1(2)}&\tiny{1.75E-1(4)}&\tiny{1.80E-1(3)}&\tiny{0(8)}&\tiny{2.29E-2(7)}\\\cline{3-12} 
&8&\tiny{1.49E-1(6)}&\tiny{2.20E-1(5)}&\tiny{-(9.5)}&\tiny{-(9.5)}&\tiny{\textbf{4.75E-1(1)}}&\tiny{4.06E-1(2)}&\tiny{2.31E-1(4)}&\tiny{2.60E-1(3)}&\tiny{0(7.5)}&\tiny{0(7.5)}\\\cline{3-12} 
&10&\tiny{1.77E-1(6)}&\tiny{2.19E-1(5)}&\tiny{-(9.5)}&\tiny{-(9.5)}&\tiny{\textbf{5.00E-1(1)}}&\tiny{4.20E-1(2)}&\tiny{2.28E-1(4)}&\tiny{2.59E-1(3)}&\tiny{0(7.5)}&\tiny{0(7.5)}\\\hline 
\multicolumn{2}{c}{Avg Rank}&$7.25$&$4.88$&$9.35$&$9.35$&$\mathbf{1.17}$&$2.33$&$4.67$&$4.12$&$6.06$&$5.81$\\ \hline 
\end{tabular}
 \label{tab:uniformLevel100000}\end{table*}

From the runtime results in Table \ref{tab:time100000}, we can see that RVSS-AD and RVSS-PD are the fastest, followed by IDSS. These methods can be executed within two seconds. Therefore, they are preferable to use in practice with high efficiency demand for large-scale subset selection. GAHSS, DSS, and CSS-MEA have medial runtime. They can be executed within 100 seconds for all the examined candidate solution sets of size 100K. Therefore, they are preferable to use in the situation where the efficiency is not so highly demanded. The other methods need much more time especially GIGDSS and GIGD+SS cannot return the result within one hour, which means that they need some modification to decrease the computation time in their applications to large-scale subset selection. When the size of candidate solution sets is small, GIGDSS and GIGS+SS show good performance with respect to the IGD and IGD+ indicators, respectively.
\begin{table*}[!tbp]
\fontsize{8.0pt}{0.4\baselineskip}\selectfont
\setlength{\tabcolsep}{0.8pt}
\caption{The runtime of each subset selection method on each candidate solution set. The number of solutions in each candidate solution set is 100,000. The number in the parenthesis is the rank of the corresponding method among the 10 methods, where a smaller value indicates a better rank.}
\centering      \addtolength{\leftskip} {-2cm} \addtolength{\rightskip}{-2cm} 
       \begin{tabular}{cccccccccccc}
 \hline \multicolumn{2}{c}{\makecell{Candidate\\ Solution Set}}&\;\;\;\;GHSS\;\;\;\;&GAHSS&GIGDSS\;&GIGD+SS&\;\;\;\;\;DSS\;\;\;\;\;&IDSS&CSS-MEA\;&CSS-MED\;&RVSS-PD\;&RVSS-AD\\ \hline 
\multirow{4}{*}{\makecell{Linear\\Triangular}}&3&\tiny{1.44E+2(7)}&\tiny{2.57E+1(5)}&\tiny{-(9.5)}&\tiny{-(9.5)}&\tiny{7.12E+0(4)}&\tiny{1.31E+0(3)}&\tiny{3.82E+1(6)}&\tiny{8.00E+2(8)}&\tiny{3.54E-1(2)}&\tiny{\textbf{2.03E-1(1)}}\\\cline{3-12} 
&5&\tiny{4.65E+2(7)}&\tiny{8.93E+1(6)}&\tiny{-(9.5)}&\tiny{-(9.5)}&\tiny{4.48E+1(4)}&\tiny{1.65E+0(3)}&\tiny{8.31E+1(5)}&\tiny{8.29E+2(8)}&\tiny{1.42E+0(2)}&\tiny{\textbf{2.77E-1(1)}}\\\cline{3-12} 
&8&\tiny{-(9)}&\tiny{5.97E+1(6)}&\tiny{-(9)}&\tiny{-(9)}&\tiny{3.06E+1(4)}&\tiny{1.41E+0(3)}&\tiny{3.95E+1(5)}&\tiny{6.06E+2(7)}&\tiny{5.74E-1(2)}&\tiny{\textbf{2.32E-1(1)}}\\\cline{3-12} 
&10&\tiny{-(9)}&\tiny{8.88E+1(6)}&\tiny{-(9)}&\tiny{-(9)}&\tiny{5.26E+1(5)}&\tiny{1.92E+0(3)}&\tiny{3.51E+1(4)}&\tiny{5.18E+2(7)}&\tiny{5.23E-1(2)}&\tiny{\textbf{2.31E-1(1)}}\\\hline 
\multirow{4}{*}{\makecell{Linear\\Inverted\\ Triangular}}&3&\tiny{1.11E+2(7)}&\tiny{2.50E+1(5)}&\tiny{-(9.5)}&\tiny{-(9.5)}&\tiny{5.58E+0(4)}&\tiny{1.30E+0(3)}&\tiny{4.81E+1(6)}&\tiny{7.96E+2(8)}&\tiny{4.29E-1(2)}&\tiny{\textbf{1.88E-1(1)}}\\\cline{3-12} 
&5&\tiny{2.95E+2(7)}&\tiny{8.67E+1(6)}&\tiny{-(9.5)}&\tiny{-(9.5)}&\tiny{5.60E+1(4)}&\tiny{1.62E+0(3)}&\tiny{7.78E+1(5)}&\tiny{8.46E+2(8)}&\tiny{1.28E+0(2)}&\tiny{\textbf{6.30E-1(1)}}\\\cline{3-12} 
&8&\tiny{6.99E+2(8)}&\tiny{5.76E+1(6)}&\tiny{-(9.5)}&\tiny{-(9.5)}&\tiny{3.11E+1(4)}&\tiny{1.34E+0(3)}&\tiny{3.79E+1(5)}&\tiny{5.78E+2(7)}&\tiny{5.75E-1(2)}&\tiny{\textbf{3.70E-1(1)}}\\\cline{3-12} 
&10&\tiny{-(9)}&\tiny{1.02E+2(6)}&\tiny{-(9)}&\tiny{-(9)}&\tiny{7.20E+1(5)}&\tiny{1.89E+0(3)}&\tiny{3.45E+1(4)}&\tiny{4.80E+2(7)}&\tiny{4.22E-1(2)}&\tiny{\textbf{1.77E-1(1)}}\\\hline 
\multirow{4}{*}{\makecell{Concave\\Triangular}}&3&\tiny{1.37E+2(7)}&\tiny{2.59E+1(5)}&\tiny{-(9.5)}&\tiny{-(9.5)}&\tiny{7.12E+0(4)}&\tiny{1.26E+0(3)}&\tiny{3.34E+1(6)}&\tiny{7.95E+2(8)}&\tiny{6.28E-1(2)}&\tiny{\textbf{1.34E-1(1)}}\\\cline{3-12} 
&5&\tiny{3.29E+2(7)}&\tiny{1.03E+2(6)}&\tiny{-(9.5)}&\tiny{-(9.5)}&\tiny{6.03E+1(4)}&\tiny{1.94E+0(3)}&\tiny{6.86E+1(5)}&\tiny{8.30E+2(8)}&\tiny{8.42E-1(2)}&\tiny{\textbf{3.48E-1(1)}}\\\cline{3-12} 
&8&\tiny{1.71E+3(8)}&\tiny{5.31E+1(5)}&\tiny{-(9.5)}&\tiny{-(9.5)}&\tiny{3.46E+1(4)}&\tiny{1.34E+0(3)}&\tiny{6.07E+1(6)}&\tiny{6.54E+2(7)}&\tiny{6.81E-1(2)}&\tiny{\textbf{4.99E-1(1)}}\\\cline{3-12} 
&10&\tiny{-(9)}&\tiny{7.82E+1(5)}&\tiny{-(9)}&\tiny{-(9)}&\tiny{8.08E+1(6)}&\tiny{1.79E+0(3)}&\tiny{7.13E+1(4)}&\tiny{5.63E+2(7)}&\tiny{5.65E-1(2)}&\tiny{\textbf{2.14E-1(1)}}\\\hline 
\multirow{4}{*}{\makecell{Concave\\Inverted\\ Triangular}}&3&\tiny{8.38E+1(7)}&\tiny{2.54E+1(5)}&\tiny{-(9.5)}&\tiny{-(9.5)}&\tiny{5.59E+0(4)}&\tiny{1.31E+0(3)}&\tiny{4.06E+1(6)}&\tiny{7.81E+2(8)}&\tiny{5.92E-1(2)}&\tiny{\textbf{1.40E-1(1)}}\\\cline{3-12} 
&5&\tiny{1.78E+2(7)}&\tiny{1.15E+2(6)}&\tiny{-(9.5)}&\tiny{-(9.5)}&\tiny{6.06E+1(5)}&\tiny{1.69E+0(3)}&\tiny{5.99E+1(4)}&\tiny{8.33E+2(8)}&\tiny{6.80E-1(2)}&\tiny{\textbf{4.42E-1(1)}}\\\cline{3-12} 
&8&\tiny{2.39E+2(7)}&\tiny{6.39E+1(6)}&\tiny{-(9.5)}&\tiny{-(9.5)}&\tiny{2.91E+1(4)}&\tiny{1.41E+0(3)}&\tiny{4.79E+1(5)}&\tiny{6.49E+2(8)}&\tiny{7.02E-1(2)}&\tiny{\textbf{4.75E-1(1)}}\\\cline{3-12} 
&10&\tiny{-(9)}&\tiny{1.00E+2(6)}&\tiny{-(9)}&\tiny{-(9)}&\tiny{8.02E+1(5)}&\tiny{1.84E+0(3)}&\tiny{4.40E+1(4)}&\tiny{5.39E+2(7)}&\tiny{4.02E-1(2)}&\tiny{\textbf{3.23E-1(1)}}\\\hline 
\multirow{4}{*}{\makecell{Convex\\Triangular}}&3&\tiny{1.54E+2(7)}&\tiny{2.55E+1(5)}&\tiny{-(9.5)}&\tiny{-(9.5)}&\tiny{7.03E+0(4)}&\tiny{1.27E+0(3)}&\tiny{4.07E+1(6)}&\tiny{8.01E+2(8)}&\tiny{3.40E-1(2)}&\tiny{\textbf{1.86E-1(1)}}\\\cline{3-12} 
&5&\tiny{5.67E+2(7)}&\tiny{9.16E+1(6)}&\tiny{-(9.5)}&\tiny{-(9.5)}&\tiny{4.16E+1(4)}&\tiny{1.95E+0(2)}&\tiny{7.10E+1(5)}&\tiny{8.21E+2(8)}&\tiny{2.11E+0(3)}&\tiny{\textbf{7.23E-1(1)}}\\\cline{3-12} 
&8&\tiny{-(9)}&\tiny{5.09E+1(6)}&\tiny{-(9)}&\tiny{-(9)}&\tiny{2.63E+1(4)}&\tiny{1.41E+0(3)}&\tiny{4.96E+1(5)}&\tiny{6.29E+2(7)}&\tiny{1.26E+0(2)}&\tiny{\textbf{3.09E-1(1)}}\\\cline{3-12} 
&10&\tiny{-(9)}&\tiny{8.27E+1(6)}&\tiny{-(9)}&\tiny{-(9)}&\tiny{8.17E+1(5)}&\tiny{1.79E+0(3)}&\tiny{4.13E+1(4)}&\tiny{5.34E+2(7)}&\tiny{8.11E-1(2)}&\tiny{\textbf{1.31E-1(1)}}\\\hline 
\multirow{4}{*}{\makecell{Convex\\Inverted\\ Triangular}}&3&\tiny{1.05E+2(7)}&\tiny{2.49E+1(5)}&\tiny{-(9.5)}&\tiny{-(9.5)}&\tiny{5.58E+0(4)}&\tiny{1.28E+0(3)}&\tiny{3.77E+1(6)}&\tiny{7.92E+2(8)}&\tiny{5.56E-1(2)}&\tiny{\textbf{1.07E-1(1)}}\\\cline{3-12} 
&5&\tiny{2.55E+2(7)}&\tiny{9.41E+1(6)}&\tiny{-(9.5)}&\tiny{-(9.5)}&\tiny{5.47E+1(4)}&\tiny{2.10E+0(3)}&\tiny{7.27E+1(5)}&\tiny{8.12E+2(8)}&\tiny{1.08E+0(2)}&\tiny{\textbf{4.94E-1(1)}}\\\cline{3-12} 
&8&\tiny{1.73E+2(7)}&\tiny{5.55E+1(5)}&\tiny{-(9.5)}&\tiny{-(9.5)}&\tiny{2.85E+1(4)}&\tiny{1.39E+0(3)}&\tiny{6.87E+1(6)}&\tiny{6.54E+2(8)}&\tiny{5.80E-1(2)}&\tiny{\textbf{4.80E-1(1)}}\\\cline{3-12} 
&10&\tiny{8.04E+2(8)}&\tiny{8.69E+1(6)}&\tiny{-(9.5)}&\tiny{-(9.5)}&\tiny{8.41E+1(5)}&\tiny{1.80E+0(3)}&\tiny{5.90E+1(4)}&\tiny{5.77E+2(7)}&\tiny{4.28E-1(2)}&\tiny{\textbf{3.42E-1(1)}}\\\hline 
\multicolumn{2}{c}{Avg Rank}&$7.71$&$5.62$&$9.35$&$9.35$&$4.33$&$2.96$&$5.04$&$7.58$&$2.04$&$\mathbf{1.00}$\\ \hline 
\end{tabular}
 \label{tab:time100000}\end{table*}

In Table \ref{tab:indicators100000}, we summarize the rank of the 10 subset selection methods with respect to different performance metrics. Based on the results, we can see that different subset selection methods have different preference on the performance metrics. This can help EMO researchers to choose a proper subset selection method for their own goals of subset selection.
\begin{table*}[!tbp]
\fontsize{8.0pt}{0.5\baselineskip}\selectfont
\setlength{\tabcolsep}{0.8pt}
\centering     
\caption{A summary of the rank of the 10 subset selection methods with respect to different performance metrics.}
\addtolength{\leftskip} {-2cm} 
\addtolength{\rightskip}{-2cm} 
\begin{tabular}{ccccccccccc}
 \hline Performance Metric\;&GHSS\;&GAHSS\;&GIGDSS\;&GIGD+SS\;&DSS\;&IDSS\;&CSS-MEA\;&CSS-MED\;&RVSS-PD\;&RVSS-AD\\ \hline 
Hypervolume&$3.42$&$\mathbf{2.54}$&$9.35$&$9.35$&$3.42$&$4.67$&$6.04$&$5.79$&$5.38$&$5.04$\\ \hline 
IGD&$6.83$&$4.83$&$9.35$&$9.35$&$5.12$&$3.33$&$\mathbf{1.33}$&$1.67$&$6.54$&$6.62$\\ \hline 
IGD+&$4.92$&$\mathbf{3.29}$&$9.35$&$9.35$&$4.75$&$4.25$&$3.50$&$3.33$&$6.00$&$6.25$\\ \hline 
Uniformity level&$7.25$&$4.88$&$9.35$&$9.35$&$\mathbf{1.17}$&$2.33$&$4.67$&$4.12$&$6.06$&$5.81$\\ \hline 
Runtime&$7.71$&$5.62$&$9.35$&$9.35$&$4.33$&$2.96$&$5.04$&$7.58$&$2.04$&$\mathbf{1.00}$\\ \hline 
\end{tabular}
\label{tab:indicators100000}\end{table*}

\subsubsection{Candidate solution sets from EMO algorithms}
We also  run the 10 subset selection methods on the candidate solution sets from the EMO algorithms. All the experimental results and analysis are provided in Section III of the supplementary material and \url{https://github.com/HisaoLabSUSTC/BenchSS}. Our experimental results on the candidate solution sets from the EMO algorithms and their analysis are similar to the case of candidate solution sets on Pareto fronts. 

\section{Conclusions}
In this paper, we proposed a benchmark test suite for large-scale subset selection, which contains a number of ready-to-use non-dominated candidate solution sets. Some candidate solution sets in our test suite were created by directly sampling points on Pareto fronts with regular and irregular shapes. The other candidate solution sets were created by running representative EMO algorithms on popular multi-objective test problems. We also conducted benchmarking studies on 10 subset selection methods using the proposed test suite. Some observations which can be guidelines for practitioners are as follows:
\begin{itemize}
\item Better solution sets than the final population are usually obtained by subset selection from non-dominated solutions among the examined solutions. 
\item The best subset selection method for a specific performance metric is the one whose objective is to maximize/minimize this performance metric. Thus, the choice of a subset selection method  should depend on the decision-maker's preference on the performance metrics. 
\item The fastest subset selection methods are the reference vector-based methods. However, the quality of the selected subset strongly depends on the shape of the distribution of solutions in the candidate solution set. It is suggested to use when the candidate solutions are distributed in  a triangular shape.
\item The GHSS method has the best hypervolume performance. However, its runtime is also long especially for the case of many-objective problems. The GAHSS method is much faster than GHSS and has the second best hypervolume performance. Thus, GAHSS is suggested to use in practice to replace GHSS when the computation time of GHSS is impractical.
\item  The GIGDSS and GIGD+SS methods are the slowest especially for large candidate solution sets. Thus, they are not suggested to use in practice when the candidate solution set is large.
\item The CSS-MEA method is very fast. It also has a good IGD performance. Thus, it is recommended to use in practice to replace GIGDSS when the computation time of GIGDSS is impractical. 
\item The DSS and IDSS methods are very fast. They have the best uniformity level performance and medial performance with respect to the other performance metrics. Thus, they are suggested to use in practice if a uniform subset is preferred. 
\end{itemize}

We did not cover all the subset selection methods in the EMO field in our benchmarking studies. One important future research topic is to have a comprehensive survey on the subset selection methods and identify their explicit or implicit objective functions. Then, we can compare the surveyed methods using the proposed test suite. Another important future research direction is to decrease the computation time of each method for large candidate solution sets (especially for the subset selection methods based on IGD and IGD+). It is also possible to extend our test suite by considering more types of Pareto front shapes, and more EMO algorithms running on more types of test problems. We hope that the proposed benchmark test suite will help to promote further progress of the subset selection research in the EMO field. 

All source codes, data sets, and experimental results in this paper are available at \url{https://github.com/HisaoLabSUSTC/BenchSS}.


\bibliographystyle{elsarticle-num}
\bibliography{sample-bibliography} 

\begin{thebibliography}{10}
\expandafter\ifx\csname url\endcsname\relax
  \def\url#1{\texttt{#1}}\fi
\expandafter\ifx\csname urlprefix\endcsname\relax\def\urlprefix{URL }\fi
\expandafter\ifx\csname href\endcsname\relax
  \def\href#1#2{#2} \def\path#1{#1}\fi

\bibitem{qian2019distributed}
C.~Qian, Distributed {Pareto} optimization for large-scale noisy subset
  selection, IEEE Transactions on Evolutionary Computation 24~(4) (2020)
  694--707.

\bibitem{kempe2003maximizing}
D.~Kempe, J.~Kleinberg, {\'E}.~Tardos, Maximizing the spread of influence
  through a social network, in: Proceedings of the 9th ACM SIGKDD International
  Conference on Knowledge Discovery and Data Mining, 2003, pp. 137--146.

\bibitem{miller2002subset}
A.~Miller, Subset selection in regression, CRC Press, 2002.

\bibitem{farahat2011efficient}
A.~K. Farahat, A.~Ghodsi, M.~S. Kamel, An efficient greedy method for
  unsupervised feature selection, in: 2011 IEEE 11th International Conference
  on Data Mining, IEEE, 2011, pp. 161--170.

\bibitem{krause2008near}
A.~Krause, A.~Singh, C.~Guestrin, Near-optimal sensor placements in gaussian
  processes: Theory, efficient algorithms and empirical studies., Journal of
  Machine Learning Research 9~(2).

\bibitem{feige1998threshold}
U.~Feige, A threshold of ln n for approximating set cover, Journal of the ACM
  (JACM) 45~(4) (1998) 634--652.

\bibitem{dueck2007non}
D.~Dueck, B.~J. Frey, Non-metric affinity propagation for unsupervised image
  categorization, in: 2007 IEEE 11th International Conference on Computer
  Vision, IEEE, 2007, pp. 1--8.

\bibitem{bringmann2014two}
K.~Bringmann, T.~Friedrich, P.~Klitzke, Two-dimensional subset selection for
  hypervolume and epsilon-indicator, in: Proceedings of the 2014 Annual
  Conference on Genetic and Evolutionary Computation, ACM, 2014, pp. 589--596.

\bibitem{kuhn2016hypervolume}
T.~Kuhn, C.~M. Fonseca, L.~Paquete, S.~Ruzika, M.~M. Duarte, J.~R. Figueira,
  Hypervolume subset selection in two dimensions: Formulations and algorithms,
  Evolutionary Computation 24~(3) (2016) 411--425.

\bibitem{guerreiro2016greedy}
A.~P. Guerreiro, C.~M. Fonseca, L.~Paquete, Greedy hypervolume subset selection
  in low dimensions, Evolutionary Computation 24~(3) (2016) 521--544.

\bibitem{zitzler2003performance}
E.~Zitzler, L.~Thiele, M.~Laumanns, C.~Fonseca, V.~da~Fonseca, Performance
  assessment of multiobjective optimizers: an analysis and review, IEEE
  Transactions on Evolutionary Computation 7~(2) (2003) 117--132.

\bibitem{singh2018distance}
H.~K. Singh, K.~S. Bhattacharjee, T.~Ray, Distance-based subset selection for
  benchmarking in evolutionary multi/many-objective optimization, IEEE
  Transactions on Evolutionary Computation 23~(5) (2019) 904--912.

\bibitem{shang2021distance}
K.~Shang, H.~Ishibuchi, Y.~Nan, Distance-based subset selection revisited, in:
  Proceedings of the Genetic and Evolutionary Computation Conference, 2021, pp.
  439--447.

\bibitem{chen2021cluster}
W.~Chen, , H.~Ishibuchi, K.~Shang, Clustering-based subset selection in
  evolutionary multiobjective optimization, in: IEEE International Conference
  on Systems, Man, and Cybernetics (SMC), 2021.
  http://arxiv.org/abs/2108.08453.

\bibitem{deb2013evolutionary}
K.~Deb, H.~Jain, An evolutionary many-objective optimization algorithm using
  reference-point-based nondominated sorting approach, {P}art {I}: {S}olving
  problems with box constraints, IEEE Transactions on Evolutionary Computation
  18~(4) (2014) 577--601.

\bibitem{coello2004study}
C.~A.~C. Coello, M.~R. Sierra, A study of the parallelization of a
  coevolutionary multi-objective evolutionary algorithm, in: Mexican
  International Conference on Artificial Intelligence, Springer, 2004, pp.
  688--697.

\bibitem{ishibuchi2015modified}
H.~Ishibuchi, H.~Masuda, Y.~Tanigaki, Y.~Nojima, Modified distance calculation
  in generational distance and inverted generational distance, in:
  International Conference on Evolutionary Multi-Criterion Optimization,
  Springer, 2015, pp. 110--125.

\bibitem{ishibuchi2020new}
H.~Ishibuchi, L.~M. Pang, K.~Shang, A new framework of evolutionary
  multi-objective algorithms with an unbounded external archive, in: ECAI 2020,
  IOS Press, 2020, pp. 283--290.

\bibitem{li2019empirical}
M.~Li, X.~Yao, An empirical investigation of the optimality and monotonicity
  properties of multiobjective archiving methods, in: International Conference
  on Evolutionary Multi-Criterion Optimization, Springer, 2019, pp. 15--26.

\bibitem{deb2002fast}
K.~Deb, A.~Pratap, S.~Agarwal, T.~Meyarivan, A fast and elitist multiobjective
  genetic algorithm: {NSGA-II}, IEEE Transactions on Evolutionary Computation
  6~(2) (2002) 182--197.

\bibitem{koppen2007substitute}
M.~K{\"o}ppen, K.~Yoshida, Substitute distance assignments in {NSGA-II} for
  handling many-objective optimization problems, in: International Conference
  on Evolutionary Multi-Criterion Optimization, Springer, 2007, pp. 727--741.

\bibitem{zitzler2001spea2}
E.~Zitzler, M.~Laumanns, L.~Thiele, {SPEA2}: Improving the strength {Pareto}
  evolutionary algorithm, TIK-report 103.

\bibitem{li2013shift}
M.~Li, S.~Yang, X.~Liu, Shift-based density estimation for {Pareto-based}
  algorithms in many-objective optimization, IEEE Transactions on Evolutionary
  Computation 18~(3) (2014) 348--365.

\bibitem{zitzler2004indicator}
E.~Zitzler, S.~K{\"u}nzli, Indicator-based selection in multiobjective search,
  in: International Conference on Parallel Problem Solving from Nature,
  Springer, 2004, pp. 832--842.

\bibitem{beume2007sms}
N.~Beume, B.~Naujoks, M.~Emmerich, {SMS-EMOA}: Multiobjective selection based
  on dominated hypervolume, European Journal of Operational Research 181~(3)
  (2007) 1653--1669.

\bibitem{zhang2007moea}
Q.~Zhang, H.~Li, {MOEA/D}: A multiobjective evolutionary algorithm based on
  decomposition, IEEE Transactions on Evolutionary Computation 11~(6) (2007)
  712--731.

\bibitem{qiu2021evolutionary}
W.~Qiu, J.~Zhu, G.~Wu, M.~Fan, P.~N. Suganthan, Evolutionary many-objective
  algorithm based on fractional dominance relation and improved objective space
  decomposition strategy, Swarm and Evolutionary Computation 60 (2021) 100776.

\bibitem{liu2020self}
S.~Liu, Q.~Lin, K.-C. Wong, C.~A.~C. Coello, J.~Li, Z.~Ming, J.~Zhang, A
  self-guided reference vector strategy for many-objective optimization, IEEE
  Transactions on Cybernetics.

\bibitem{zhu2021new}
S.~Zhu, L.~Xu, E.~D. Goodman, Z.~Lu, A new many-objective evolutionary
  algorithm based on generalized {Pareto} dominance, IEEE Transactions on
  Cybernetics.

\bibitem{li2018two}
K.~Li, R.~Chen, G.~Fu, X.~Yao, Two-archive evolutionary algorithm for
  constrained multiobjective optimization, IEEE Transactions on Evolutionary
  Computation 23~(2) (2019) 303--315.

\bibitem{zhang2020enhancing}
Y.~Zhang, G.-G. Wang, K.~Li, W.-C. Yeh, M.~Jian, J.~Dong, Enhancing {MOEA/D}
  with information feedback models for large-scale many-objective optimization,
  Information Sciences 522 (2020) 1--16.

\bibitem{zhang2016decision}
X.~Zhang, Y.~Tian, R.~Cheng, Y.~Jin, A decision variable clustering-based
  evolutionary algorithm for large-scale many-objective optimization, IEEE
  Transactions on Evolutionary Computation 22~(1) (2018) 97--112.

\bibitem{jiang2015simple}
S.~Jiang, J.~Zhang, Y.-S. Ong, A.~N. Zhang, P.~S. Tan, A simple and fast
  hypervolume indicator-based multiobjective evolutionary algorithm, IEEE
  Transactions on Cybernetics 45~(10) (2015) 2202--2213.

\bibitem{bringmann2017maximum}
K.~Bringmann, S.~Cabello, M.~Emmerich, Maximum volume subset selection for
  anchored boxes, in: 33rd International Symposium on Computational Geometry,
  Schloss Dagstuhl, 2017, pp. 1--15.

\bibitem{shang2020survey}
K.~Shang, H.~Ishibuchi, L.~He, L.~M. Pang, A survey on the hypervolume
  indicator in evolutionary multiobjective optimization, IEEE Transactions on
  Evolutionary Computation 25~(1) (2021) 1--20.

\bibitem{bradstreet2007incrementally}
L.~Bradstreet, L.~While, L.~Barone, Incrementally maximising hypervolume for
  selection in multi-objective evolutionary algorithms, in: 2007 IEEE Congress
  on Evolutionary Computation, IEEE, 2007, pp. 3203--3210.

\bibitem{ishibuchi2009selecting}
H.~Ishibuchi, Y.~Sakane, N.~Tsukamoto, Y.~Nojima, Selecting a small number of
  representative non-dominated solutions by a hypervolume-based solution
  selection approach, in: 2009 IEEE International Conference on Fuzzy Systems,
  IEEE, 2009, pp. 1609--1614.

\bibitem{friedrich2014maximizing}
T.~Friedrich, F.~Neumann, Maximizing submodular functions under matroid
  constraints by multi-objective evolutionary algorithms, in: International
  Conference on Parallel Problem Solving from Nature, Springer, 2014, pp.
  922--931.

\bibitem{sayin2000measuring}
S.~Say{\i}n, Measuring the quality of discrete representations of efficient
  sets in multiple objective mathematical programming, Mathematical Programming
  87~(3) (2000) 543--560.

\bibitem{daszykowski2002representative}
M.~Daszykowski, B.~Walczak, D.~Massart, Representative subset selection,
  Analytica Chimica Acta 468~(1) (2002) 91--103.

\bibitem{das1998normal}
I.~Das, J.~E. Dennis, Normal-boundary intersection: A new method for generating
  the {Pareto} surface in nonlinear multicriteria optimization problems, SIAM
  Journal on Optimization 8~(3) (1998) 631--657.

\bibitem{ishibuchi2017performance}
H.~Ishibuchi, Y.~Setoguchi, H.~Masuda, Y.~Nojima, Performance of
  decomposition-based many-objective algorithms strongly depends on {Pareto}
  front shapes, IEEE Transactions on Evolutionary Computation 21~(2) (2017)
  169--190.

\bibitem{ulrich2012bounding}
T.~Ulrich, L.~Thiele, Bounding the effectiveness of hypervolume-based ($\mu$+
  $\lambda$)-archiving algorithms, in: International Conference on Learning and
  Intelligent Optimization, Springer, 2012, pp. 235--249.

\bibitem{chen2021fast}
W.~Chen, H.~Ishibuchi, K.~Shang, Fast greedy subset selection from large
  candidate solution sets in evolutionary multi-objective optimization, IEEE
  Transactions on Evolutionary Computation.

\bibitem{nemhauser1978analysis}
G.~L. Nemhauser, L.~A. Wolsey, M.~L. Fisher, An analysis of approximations for
  maximizing submodular set functions—{I}, Mathematical Programming 14~(1)
  (1978) 265--294.

\bibitem{ahmadi2019uniform}
A.~Ahmadi-Javid, A.~Moeini, Uniform distributions and random variate generation
  over generalized lp balls and spheres, Journal of Statistical Planning and
  Inference 201 (2019) 1--19.

\bibitem{deb2005scalable}
K.~Deb, L.~Thiele, M.~Laumanns, E.~Zitzler, Scalable test problems for
  evolutionary multiobjective optimization, in: Evolutionary Multiobjective
  Optimization, Springer, 2005, pp. 105--145.

\bibitem{huband2006review}
S.~Huband, P.~Hingston, L.~Barone, L.~While, A review of multiobjective test
  problems and a scalable test problem toolkit, IEEE Transactions on
  Evolutionary Computation 10~(5) (2006) 477--506.

\bibitem{While2012A}
L.~While, L.~Bradstreet, L.~Barone, A fast way of calculating exact
  hypervolumes, IEEE Transactions on Evolutionary Computation 16~(1) (2012)
  86--95.

\bibitem{shang2021greedy}
K.~Shang, H.~Ishibuchi, W.~Chen, Greedy approximated hypervolume subset
  selection for many-objective optimization, in: Proceedings of the Genetic and
  Evolutionary Computation Conference, 2021, pp. 448--456.

\bibitem{shang2018r2}
K.~Shang, H.~Ishibuchi, X.~Ni, R2-based hypervolume contribution approximation,
  IEEE Transactions on Evolutionary Computation 24~(1) (2020) 185--192.

\bibitem{tian2017platemo}
Y.~Tian, R.~Cheng, X.~Zhang, Y.~Jin, {PlatEMO}: A {MATLAB} platform for
  evolutionary multi-objective optimization [educational forum], IEEE
  Computational Intelligence Magazine 12~(4) (2017) 73--87.

\bibitem{ishibuchi2019comparison}
H.~Ishibuchi, R.~Imada, N.~Masuyama, Y.~Nojima, Comparison of hypervolume,
  {IGD} and {IGD+} from the viewpoint of optimal distributions of solutions,
  in: 2019 International Conference on Evolutionary Multi-Criterion
  Optimization, Springer, 2019, pp. 332--345.

\bibitem{ishibuchi2018reference}
H.~Ishibuchi, R.~Imada, Y.~Setoguchi, Y.~Nojima, Reference point specification
  in inverted generational distance for triangular linear {Pareto} front, IEEE
  Transactions on Evolutionary Computation 22~(6) (2018) 961--975.

\end{thebibliography}

\end{document}